%% file: main.tex
\documentclass[10pt]{article} 
\usepackage[accepted]{tmlr}

\input{math_commands.tex}

\usepackage{amsmath, amssymb, amsthm}
\newtheorem{theorem}{Theorem}
\newtheorem{lemma}{Lemma}

\newcommand{\Prob}{\mathbb{P}}
\newcommand{\norm}[1]{\left\lVert #1 \right\rVert}
\newcommand{\HALL}{\mathrm{H}_\delta} 

\usepackage{hyperref}
\usepackage{url}
\usepackage{graphicx}
\usepackage{multirow}
\usepackage{longtable}

\title{Neural Diversity Regularizes Hallucinations\\ in Language Models}

\author{\name Kushal Chakrabarti \email kushalc@obviouslywrong.org\\
  \name Nirmal Balachundhar \email nbalachundhar@gmail.com\\
  \addr South Park Commons, San Francisco, CA
}

\def\month{07}
\def\year{2026}
\def\openreview{\url{https://openreview.net/forum?id=5l9ZflyApA}}

\begin{document}

\maketitle

\begin{abstract}
  Language models continue to hallucinate despite scaling parameters, compute, and data. We propose
  \emph{neural diversity} --- decorrelated parallel representations --- as a provable mechanism to reduce
  hallucination rates at fixed parameter and data budgets. While existing mitigation strategies largely
  target accuracy, we \emph{reframe it as a second-moment reliability problem} governed by representational
  covariance and \emph{provide the first formal tail bounds} for hallucination probability in ensembled language models,
  explaining 94.3\% of reliability variation across configurations in our setting (Qwen2.5-0.5B, 20M
  Pile tokens, 12 tasks). We introduce ND-LoRA (Neural Diversity Low-Rank Adaptation), combining parallel
  LoRA adapters with Barlow Twins regularization, and \emph{reduce hallucinations by up to 25.6\% (and 14.6\%
  on average)} while preserving capability. Ablations show LoRA and regularization act synergistically;
  causal interventions identify neural diversity as the mediating factor; correlational studies indicate
  scale, with +0.1\% neural correlation associated with +3.8\% hallucination. Finally, task-dependent
  optima emerge: different tasks require different optimal diversity. Neural diversity enables
  reliability gains without scaling, improving tails orthogonally to parameters and data at near-zero
  cost (+0.008\% pretraining, 1.1$\times$ latency).
\end{abstract}

\begin{figure}[!h]
  \begin{center}
    \includegraphics[width=0.68\textwidth]{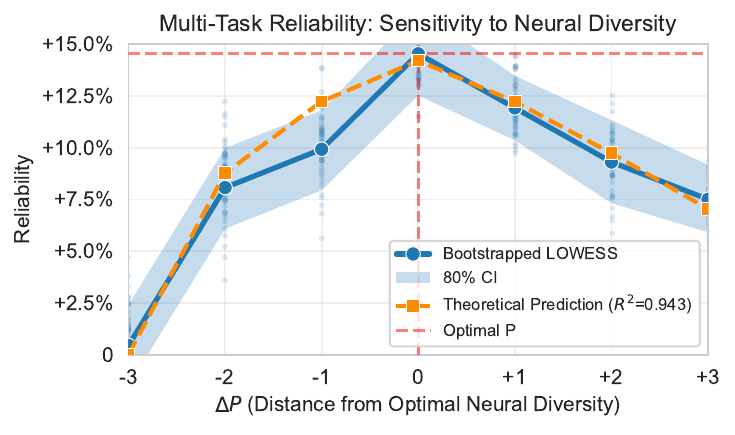}
  \end{center}
  \caption{
    \textbf{Maximizing reliability requires optimal neural diversity.}
    Across $P \in \{1,2,4,8\}$ parallel representations and 6 hallucination benchmarks (182,850 samples,
    LOWESS, 80\% CI), reliability follows an inverted-U, peaking at optimal $P_\star$ ($\Delta P = P -
    P_\star$) then degrading. Theorems \ref{thm:halluc-with-diversity} \& \ref{thm:u-shape} precisely predict the
    corresponding U-shaped hallucination probability ($R^2 = 0.943$, orange) and motivate principled
    architectural design (ND-LoRA), which reduces hallucinations 14.6\% on average without degrading general
    capabilities.
  }
  \label{fig:optimaldiv}
\end{figure}

\section{Introduction}
Despite scaling to trillions of parameters, language models hallucinate persistently \citep{lin2021truthfulqa}.
This reliability crisis is acute for small language models --- increasingly favored for edge and
agentic use cases \citep{chen2024edge, belcak2025agents} --- whose compressed representations make them
especially vulnerable to hallucinations, with even well-resourced efforts like GPT-OSS 20B exhibiting 91\%
hallucination rates on factual benchmarks \citep{openai2025gptoss}.

Current hallucination mitigation strategies are largely empirically driven but theoretically ungrounded and target
average performance rather than tail risk. RLHF optimizes mean harmlessness \citep{bai2022constitutional},
RAG improves average factual grounding \citep{niu2024ragtruth}, and contrastive decoding enhances mean
generation quality \citep{li2023contrastive}. While inference-time approaches like self-consistency and LoRA ensembling
\citep{wang2022selfconsistency, wang2023loraensemble} reduce hallucinations through diverse sampling, they lack formal
tail-probability guarantees. Similarly, parallel scaling methods \citep{chen2025parscale} target first-moment
improvements in perplexity and task accuracy. Yet controlling catastrophic failures requires bounding the tails of
$\Prob(\text{hallucination})$, not just optimizing mean behavior.

\begin{table}[t]
  \begin{center}
    \begin{tabular}{ll|cccc}
      \multicolumn{1}{c}{\bf Category} &
      \multicolumn{1}{c}{\bf Task} &
      \multicolumn{1}{c}{\bf Best $P_\star$} &
      \multicolumn{1}{c}{\bf Best Score} &
      \multicolumn{1}{c}{\bf ${\Delta}$\% Score} &
      \multicolumn{1}{c}{\bf Sig.}
      \\ \hline \\[-0.75em]
      \multirow{6}{*}{Hallucination}
      & HaluEval (Dialog)    & 4 & 0.516 & +12.8\% & *** \\
      & HaluEval (QA)        & 4 & 0.451 & +23.4\% & *** \\
      & HaluEval (Summ)      & 4 & 0.502 & +25.6\% & *** \\
      & MemoTrap v2          & 8 & 0.689 & +8.8\%  & *** \\
      & TruthfulQA (MC1)     & 2 & 0.269 & +7.3\%  & \\
      & TruthfulQA (MC2)     & 2 & 0.442 & +9.5\%  & * \\
      \hline \\[-0.75em]
      \multirow{4}{*}{Knowledge}
      & NQ (8-shot)          & 1 & 0.066 & --      & \\
      & NQ-swap              & 8 & 0.554 & +0.8\%  & \\
      & PopQA                & 1 & 0.111 & --      & \\
      & TriviaQA (8-shot)    & 1 & 0.192 & --      & \\
    \end{tabular}
  \end{center}
  \caption{
    \textbf{Optimal neural diversity is task-dependent: hallucination tasks benefit from neural diversity,
    knowledge tasks do not.} De-aggregating \autoref{fig:optimaldiv}, hallucination benchmarks consistently
    show large gains with increased diversity (up to 25.6\%, HaluEval-Summ, $P_\star = 4$), while knowledge
    retrieval mostly peaks at $P_\star = 1$. This asymmetry supports hallucination as a reliability problem
    distinct from factual recall. Significance: ***\,$p < 0.001$, *\,$p < 0.05$.
  }
  \label{tab:taskdiv}
\end{table}

Formal ensemble theory exists but targets the wrong objective. Classical ensemble methods
\citep{krogh1995neural} provide rigorous diversity theory to reduce mean generalization error
$\E[\text{loss}]$, not tail-probability bounds for hallucinations. Deep ensembles
\citep{lakshminarayanan2017deep} quantify uncertainty but lack hallucination-specific guarantees. Without
explicit diversification, parallel architectures suffer \emph{representational collapse}
\citep{jing2022dimensional}, leaving reliability gains unrealized.

Our key insight is that hallucinations --- to the extent they arise from correlated representational errors
rather than missing knowledge --- are a form of noise amenable to portfolio-theoretic diversification. We
demonstrate that a significant subset of hallucinations are in fact empirically addressable and
theoretically tail-boundable by such diversification.

To our knowledge, we provide the first formal framework for \textbf{hallucination probability tail bounds in
ensembled language models}, reframing it as a second-moment reliability problem. Drawing on portfolio theory
\citep{markowitz1952portfolio}, we prove that decorrelated parallel representations (\emph{neural diversity})
reduce this tail bound and introduce \textbf{ND-LoRA (Neural Diversity Low-Rank
Adaptation)}\footnote{Code, training and evaluation scripts, and checkpoints:
\url{https://github.com/kushalc/nd-lora}.} to concretely
demonstrate its hallucination reduction capabilities.

Our contributions are:
\begin{itemize}
  \item \textbf{Theoretical Linkage}: We reframe hallucinations as a second-moment reliability problem and prove
    (i) a portfolio-theoretic bound showing hallucination probability $\Prob(\mathrm{H})~\propto~1/P$
    with $P$ decorrelated parallel representations (\autoref{thm:halluc-with-diversity}); and, (ii)
    non-monotonicity in reliability scaling (\autoref{thm:u-shape}), showing that excessive parallelism can
    degrade diversity (and thus reliability) under common circumstances. We further show (iii) our
    theoretical predictions achieve $R^2 = 0.943$ in fitting empirical reliability gains
    (\autoref{fig:optimaldiv}), establishing quantitative validation rare in neural hallucination research.
  \item \textbf{Constructive Demonstration}: We demonstrate empirical feasibility via ND-LoRA (parallel LoRA
    + Barlow Twins decorrelation), reducing hallucinations by up to 25.6\% (and 14.6\% on average) at
    1.00008$\times$ continued pretraining cost (1.1$\times$ inference latency) while preserving general
    capabilities across 12 tasks on a small language model (Qwen2.5-0.5B) (\autoref{tab:taskdiv},
    \ref{tab:relacc}).
  \item \textbf{Mechanistic Analysis}: We establish that neural diversity mediates hallucination in four
    ways: (i) causality via perturbation ($p < 0.001$, \autoref{tab:causality}), (ii) quantitative scale via
    correlation (+0.1\% diversity $\Leftrightarrow$ -3.8\% hallucination, \autoref{fig:correlational}),
    (iii) super-linear effects via ablation (\autoref{tab:ablations}), and (iv) task-dependent optima via
    scaling sweeps (\autoref{tab:taskdiv}).
\end{itemize}

Neural diversity enables reliability gains without scaling compute. While traditional scaling asks
``how big?'' and data scaling ``how much?'', diversity asks ``how different?'' --- controlling tail
probability through variance and correlation structure rather than capacity. Because the backbone stays
frozen and only the parallel adapters train, this costs +0.008\% of continued-pretraining compute; the sole
recurring tradeoff is a 1.1$\times$ inference latency from the $P$ parallel streams (\autoref{sec:cost_analysis}).

\section{A Theory of Neural Diversity}
\label{sec:theory}
Why don't existing scaling methods improve reliability? Without explicit diversity mechanisms, gradient descent
drives parallel streams toward similar representations through \emph{representational collapse}
\citep{jing2022dimensional}, leaving reliability gains unrealized. We establish the first hallucination tail
bounds for ensembled language models, proving that neural diversity reduces hallucinations and providing
mathematical foundations for ND-LoRA.

Our strategy adapts portfolio theory to neural architecture design. Classical ensemble methods reduce
\emph{mean error} $\E[\text{loss}]$ through variance reduction \citep{krogh1995neural}, treating
correlation as a factor that limits accuracy gains. In contrast, portfolio theory manages \emph{tail risk}
--- rare but catastrophic failures --- by diversifying across correlated assets
\citep{markowitz1952portfolio}. We adapt the latter framework to tail bound hallucination probability
$\Prob(\text{hallucination})$, where correlation becomes the primary control variable for reliability rather
than a secondary constraint on mean performance. We borrow terminology and analytical tools from financial
econometrics throughout; \autoref{sec:glossary} provides a glossary of non-standard deep learning and
econometric terms for readers unfamiliar with either field.

\subsection{Preliminaries}
\label{sec:preliminaries}
Modern language models hallucinate by fabricating facts, generating content inconsistent with input, or
creating unsupported claims \citep{maynez2020faithfulness, ji2023survey}. While comprehensive taxonomies
exist \citep{huang2024survey}, a useful factoring separates \emph{faithfulness} (consistency with context)
from \emph{factuality} (correctness of recall). We primarily model the former via a simple signal-noise proxy
that captures the underlying reliability failure --- correlated representational errors --- while remaining
analytically tractable. However, as the two failure modes are not mutually exclusive, our technique can also
improve factuality where correlated errors (not lack of knowledge) are the limiting factor.

\paragraph{Signal-noise model.}
Let $x \in X$ be a query with oracle output $y_\star(x) \in \mathbb{R}^V$ for vocabulary size $V$ and
corresponding hidden representation $z_\star(x) \in \mathbb{R}^d$ for hidden dimension $d$. Consider an
architecture that employs $P$ parallel computational pathways called \emph{streams}, each processing the same
input $x$ through the same model but in perturbed ways. We model the hidden output of each stream as $Z_i =
z_\star + \varepsilon_i$ where $\varepsilon_i \in \mathbb{R}^d$ is centered noise with variance $\sigma_i^2 >
0$. The noise covariance $\Sigma \in \mathbb{R}^{P \times P}$ has entries $\Sigma_{ij} \triangleq \E[\langle
\varepsilon_i, \varepsilon_j\rangle]$ with pairwise correlations $\rho_{ij} \triangleq
\Sigma_{ij}/(\sigma_i\sigma_j)$ for $i \neq j$. We aggregate hidden representations via $\widehat{Z}_w =
\sum_i w_i Z_i$ with non-negative weights summing to one ($w_i \ge 0$, $\sum_i w_i = 1$). For readability, we
omit $x$ where obvious and denote the
average noise variance by $\bar{\sigma}^2 \triangleq \E[\sigma_i^2]$ and average correlation $\bar{\rho}
\triangleq \E_{i<j}[\rho_{ij}]$.

\paragraph{High-dimensional structure.}
High-dimensional representations exhibit predictable geometric regularity that we exploit for analysis.
We assume: (i) \emph{Lipschitz decoding}, where outputs $\widehat{Y}_w(x) = f(\widehat{Z}_w(x))$
and $y_\star(x) = f(z_\star(x))$ satisfy $\|f(z) - f(z')\|_2 \le L \|z - z'\|_2$ for some $L > 0$; (ii)
\emph{norm concentration}, where $\|\tilde{z}_i(x)\|^2_2 \approx d$ with small relative variance for
per-feature whitened representations $\tilde{z}_i$; and, (iii) \emph{feature alignment}, where in the readout
eigenbasis $(e_k^\top \tilde{z}_i)(e_k^\top \tilde{z}_j) \ge 0$ for all features $k$ and stream pairs $(i,j)$
--- i.e. features are self-consistent across streams and maintain polarity. Properties (i) and (ii) are
standard in high-dimensional probability \citep{vershynin2018high} and neural network analysis
\citep{fazlyab2019efficient, bartlett2017spectrally}. Property (iii) holds naturally in parallel
architectures: streams share a backbone and combine via convex aggregation, so each stream's projection onto
a readout eigenvector inherits the same sign from the shared backbone; \autoref{tab:ablations} confirms this
empirically (neural diversity index $\mathcal{D} \sim 1$; see below) for standard parallel architectures.

\paragraph{Neural representations.}
At a chosen design layer, each stream $i$ exposes a $d$-dimensional hidden representation $z_i(X)$. We whiten
per-feature to obtain $\tilde{z}_i$ with zero mean and identity covariance. For streams $i < j$, the
cross-correlation matrix is $C^{(ij)} \triangleq \E\big[\tilde{z}_i \tilde{z}_j^\top\big] \in \mathbb{R}^{d
\times d}$ whose diagonal entries measure same-feature similarity and off-diagonal entries capture
cross-feature alignment. Finally, using the widely-exploited observation that trained networks exhibit
locally linear behavior at their operating point \citep{goodfellow2015explaining, simonyan2014deep}, we
connect representations to noise via local linearity: $\xi_i \;=\; A \tilde{z}_i$ for a shared linear readout
$A \in \mathbb{R}^{V \times d}$ with finite condition number $\kappa = s_{\max}/s_{\min}$ over its
$\min(V,d)=d$ singular values.

\paragraph{Neural diversity index.}
We define a simple cosine-based index to measure cross-stream diversity:
\begin{equation}
  \label{eq:ndi}
  \mathcal{D}
  \triangleq \sqrt{\E_{i<j}\left[\frac{(\tilde{z}_i \cdot \tilde{z}_j)^2}{\|\tilde{z}_i\|^2 \|\tilde{z}_j\|^2}\right]}.
\end{equation}
Lower $\mathcal{D}$ indicates greater neural diversity: $\mathcal{D} = 0$ means all streams are perfectly
orthogonal, while $\mathcal{D} = 1$ means streams have suffered complete collapse.

\paragraph{Hallucinations.}
We define the output error as $E_{w} \triangleq \|\widehat{Y}_w(x) - y_\star(x)\|_F$, which is comparable to
metrics like TruthfulQA-MC2 \citep{lin2021truthfulqa}. For tolerance $\delta > 0$, the \emph{hallucination
event} is $\HALL \triangleq \{E_{w} \ge \delta\}$. Our goal is to bound $\Prob(\HALL)$ as a
function of neural diversity $\mathcal{D}$ across streams $P$.

\subsection{Neural Diversity Bounds Hallucination}
\label{sec:diversity_bounds}

Classical portfolio theory \citep{markowitz1952portfolio} gives the variance of an equally weighted portfolio
of $P$ assets with average variance $\bar{\sigma}^2$ and average pairwise correlation $\bar{\rho}$ as:
\begin{equation}
  \label{eq:portfolio-variance}
  \Var(Y) \;=\; \bar{\sigma}^2\!\left(\frac{1-\bar{\rho}}{P}+\bar{\rho}\right).
\end{equation}

To use this observation for hallucinations, we must first connect neuron-level representations to
portfolio-level noise correlations. Exploiting the fact that (i) our ensemble has one underlying
model with aligned neuron-level representations and (ii) our model has geometric regularity in representation
and output, the following lemma establishes this mapping:



\begin{lemma}[Average Correlation Bound]
  \label{lem:cosine-avg-corr}
  Suppose there exists a kurtosis bound $C_4 \ge 1$ such that $\E[\|\xi_i\|_2^4] \le C_4\,\sigma_i^4$ for all
  streams $i$. Then the average pairwise noise correlation $\bar{\rho}$ satisfies
  \begin{equation}
    |\bar{\rho}|
    \;\le\;
    C_\ast\,\mathcal{D},
  \end{equation}
  where $C_\ast \triangleq \sqrt{C_4}\,\kappa^2$ depends on the kurtosis bound $C_4$ and readout condition
  number $\kappa$.
\end{lemma}

\begin{proof}[Proof sketch]
  We proceed in two steps. Let $\mathcal{D}_\xi$ denote the diversity index computed over noise vectors
  $\xi_i$ (analogous to $\mathcal{D}$ but in readout space), with pairwise terms $\mathcal{D}_{\xi,ij}$.
  \emph{(1)} Spectral bounds and feature alignment imply the linear readout distorts cosines
  by at most $\kappa^2$, so $\mathcal{D}_\xi \le \kappa^2 \mathcal{D}$.
  \emph{(2)} Cauchy--Schwarz twice --- inner products to cosines, then kurtosis --- gives
  $|\rho_{ij}| \le \sqrt{C_4}\,\mathcal{D}_{\xi,ij}$; averaging pairs completes the proof.
\end{proof}

We now have a direct path to
tail-bound $P(\HALL)$ as a function of $\mathcal{D}$ and $P$. For readability, we assume uniform weights
$w_i = 1/P$ below but our approach can also be easily applied to arbitrary weights.

\begin{theorem}[Hallucination Bound with Diversity]
  \label{thm:halluc-with-diversity}
  For any tolerance $\delta > 0$, the hallucination probability $\Prob(\HALL)$ satisfies
  \begin{equation}
    \Prob(\HALL)
    \;\le\;
    \frac{
      \frac{1 - C_\ast\,\mathcal{D}}{P}
      +
      C_\ast\,\mathcal{D}
    }
    {
      \frac{1 - C_\ast\,\mathcal{D}}{P}
      +
      C_\ast\,\mathcal{D}
      + SNR
    },
  \end{equation}
  where $SNR \triangleq \delta^2 / \bar{\sigma}^2$ is the signal-to-noise ratio, $\mathcal{D}$ is the neural
  diversity index (\autoref{eq:ndi}), $C_\ast = \sqrt{C_4}\,\kappa^2$ is the readout-kurtosis constant
  (\autoref{lem:cosine-avg-corr}), and $P$ is the number of parallel streams as above.
\end{theorem}

\begin{proof}[Proof sketch]
  \autoref{lem:cosine-avg-corr} bounds $|\bar{\rho}| \le C_\ast\,\mathcal{D}$, linking noise variance to
  representational diversity. Plugging into \autoref{eq:portfolio-variance}, applying Chebyshev and
  normalizing by $\bar{\sigma}^2$ yields the stated bound.
\end{proof}

This completes the first half of our theoretical result: Neural diversity mediates hallucination probability.
With perfect de-correlation ($\bar{\rho} = 0$), hallucination probability scales as $O(1/P)$ --- more streams
reduce hallucination risk. When streams collapse ($\bar{\rho} = 1$), the bound becomes independent of $P$,
explaining why naive ensembling without diversification provides no reliability benefits.

\subsection{Non-Monotonic Scaling Behavior}
Next, we demonstrate that under common circumstances, the hallucination bound follows a U-shaped
curve --- initially decreasing with higher $P$, but starts increasing eventually. Consider the case where the
correlation itself increases with $P$, say, due to optimizer constraints:

\begin{theorem}[U-shaped Behavior]
  \label{thm:u-shape}
  Suppose $\bar{\rho}(P) = \rho_0 + \beta (P-1)^\gamma$ for constants $\rho_0 \in [0,1)$, $\beta > 0$,
  $\gamma > 0$. Define
  \begin{equation}
    v(P) \triangleq \Var(E_w)
    = \bar{\sigma}^2 \left(\frac{1 - \bar{\rho}(P)}{P} + \bar{\rho}(P)\right),
    \qquad
    \mathcal{B}(P) \triangleq \frac{v(P)}{v(P) + \delta^2}.
  \end{equation}
  Then $\mathcal{B}(P)$ is U-shaped: there exists $P_\star \ge 1$ minimizing
  $\Prob(\HALL)$, with $P_\star$ controlled by how fast $\bar{\rho}(P)$ degrades with $P$.
\end{theorem}

\begin{proof}[Proof sketch]
  The hallucination bound $\mathcal{B}(P)$ is monotonic in variance $v(P)$, so we analyze $v(P)$ directly.
  There are two competing effects: the $1/P$ term drives variance down, while growing correlation
  $\bar{\rho}(P) = \rho_0 + \beta(P-1)^\gamma$ eventually dominates. Differentiating shows $v'(P)$ changes
  sign exactly once, yielding a unique minimum $P_\star$ whose location depends $\beta, \gamma \text{ and } \rho_0$.
\end{proof}

This theorem establishes \emph{non-monotonicity} --- hallucination probability $\Prob(\HALL)$ actually
\emph{increases} for larger $P$, meaning reliability degrades. This is stronger than the well-known
diminishing returns of ensembles (where improvement slows but continues). While ensemble theory also shows
optimal size matches the number of class labels for accuracy-optimized classifiers \citep{bonab2019less}, we
prove and validate (\autoref{fig:optimaldiv}) that diversity can degrade in generative language models with
excessive parallelism under common circumstances and also harm reliability.

\subsection{Theoretical Validation}
By measuring empirical diversity $\mathcal{D}(P)$ and plugging these values into
\autoref{thm:halluc-with-diversity}'s bound, we achieve $R^2 = 0.943$ (\autoref{fig:optimaldiv}), explaining
94.3\% of empirical reliability variation. This fit uses only two free parameters ($C_\ast$, $SNR$)
shared across all tasks and observations, with $\mathcal{D}(P)$ fixed from empirical measurements.
\autoref{thm:u-shape}'s correlation growth model provides a mechanism for observed concavity:
correlation grows as $O((P-1)^\gamma)$, overwhelming the $O(1/P)$ diversification benefit. This alignment ---
rare in hallucination research where theory often lags empirics --- validates our portfolio-theoretic framework.

Together, \autoref{thm:halluc-with-diversity} and \autoref{thm:u-shape} show that (i) reducing
$\mathcal{D}$ reduces hallucinations and (ii) there exists an optimal $P_{\star}$ that minimizes
hallucinations. Next, we show how to construct an architecture and training protocol to reduce
$\mathcal{D}$ and find $P_{\star}$.

\section{ND-LoRA: A Practical Demonstration}
\label{sec:ndlora}
\begin{figure}[ht]
  \begin{center}
    \includegraphics[width=1\textwidth]{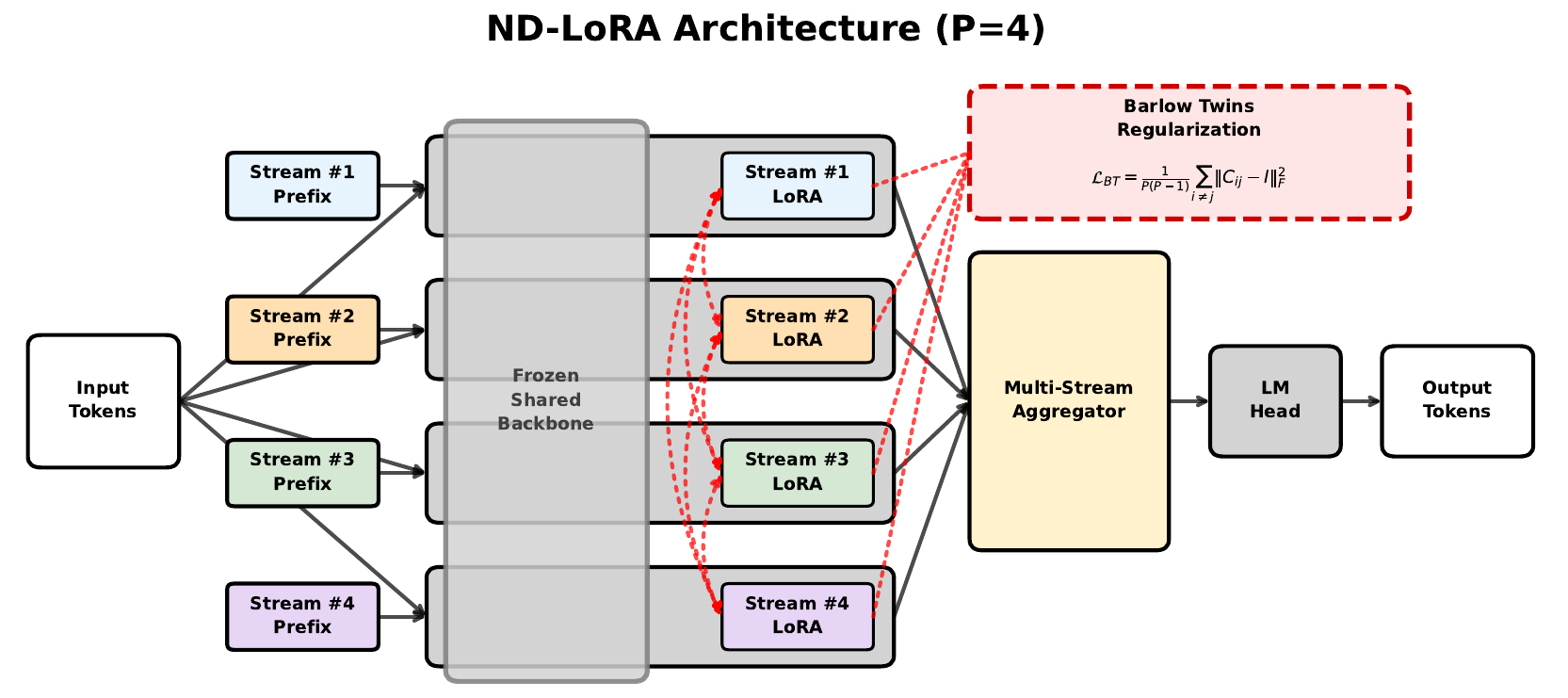}
  \end{center}
  \caption{
    \textbf{ND-LoRA schematic for $P=4$ parallel streams.} Each stream receives independent LoRA adapters
    and learnable prefix tokens. The aggregator combines stream outputs with learnable weights, while
    Barlow Twins regularization incentivizes decorrelation between stream outputs.
  }
  \label{fig:diagram}
\end{figure}

We introduce ND-LoRA (Neural Diversity Low-Rank Adaptation), a parameter-efficient method that demonstrates our
theoretical framework for neural diversity regularization. ND-LoRA extends the ParScale architecture with
stream-aware LoRA adapters and explicit decorrelation objectives. \autoref{fig:diagram} visually summarizes
our approach.

\subsection{Architecture}
Our implementation builds on ParScale with $P$ parallel computation streams. Each stream $i \in \{1,
\ldots, P\}$ uses 48 learnable prefix tokens prepended to the input sequence that flow through all layers
via the attention mechanism, along with stream-specific LoRA adapters applied at each layer:
\begin{equation}
  h_i^{(\ell)} = \text{Layer}^{(\ell)}(h_i^{(\ell-1)} + B_i^{(\ell)} A_i^{(\ell)} h_i^{(\ell-1)})
\end{equation}
where $B_i^{(\ell)} \in \mathbb{R}^{d \times r}$, $A_i^{(\ell)} \in \mathbb{R}^{r \times d}$ are stream-specific
LoRA matrices with rank $r$. The final output combines streams through a learned aggregator:
\begin{equation}
  y = \text{LM\_Head}\left(\sum_{i=1}^P w_i \cdot h_i^{(L)}\right)
\end{equation}
where $w_i = (1-\varepsilon) \cdot \text{softmax}(\text{MLP}([h_1^{(L)}, \ldots, h_P^{(L)}]))_i +
\varepsilon/P$ are dynamic weights with
label smoothing ($\varepsilon = 0.1$) computed from the concatenated stream representations.
This prevents attention collapse by ensuring minimum weight $\varepsilon/P$ for each stream.

This architecture enables stream specialization while maintaining parameter efficiency. For $P=2$ streams
with rank-16 LoRA, we use approximately 29K trainable parameters per layer, comparable to a single rank-32
LoRA but with fundamentally different representational capabilities.

Concretely, this architecture approximates \autoref{lem:cosine-avg-corr}'s shared-readout simplification. The
post-design computation factors as a frozen backbone (layers $\ell_\star{+}1, \ldots, L$) composed with a
single $\mathrm{lm\_head}$, both shared across streams. Only rank-$16$ LoRA adapters perturb this
post-design readout (pre-design adapters shape $\tilde{z}_i$ itself), giving $A_i = A + \Delta A_i$ with
$\mathrm{rank}(\Delta A_i) \le 16 \ll \min(V, d) = d = 896$, so $A_i \approx A$. The $R^2 = 0.943$
fit in \autoref{fig:optimaldiv} confirms that the theory (including modeling simplifications like
shared-readout, linearization and norm concentration) bridges well to practice.

\subsection{Barlow Twins Regularization}
To encourage neural diversity, we apply Barlow Twins regularization across all pairs of streams $i < j$ at
a pre-specified design layer $\ell_\star$.

Let $z_i \in \mathbb{R}^{B \times T \times d}$ denote the hidden representations of stream $i$ at the design
layer for a batch of size $B$ and sequence length $T$. We first apply batch normalization and mean-centering to
obtain whitened features $\tilde{z}_i$. We then calculate the cross-correlation matrices $C^{(ij)} \in
\mathbb{R}^{d \times d}$ as in \autoref{sec:diversity_bounds} and apply standard Barlow Twins
\citep{zbontar2021barlow} for each pair of streams $i < j$:
\begin{equation}
  \mathcal{L}_{BT} = \mathop{\E}_{i < j} \norm{C^{(ij)} - I}_F
\end{equation}

The total training objective combines cross-entropy and decorrelation terms:
\begin{equation}
  \mathcal{L} = \mathcal{L}_{CE} + \lambda_{BT} \mathcal{L}_{BT}
\end{equation}
Notably, driving $C^{(ij)} \to I$ reinforces the feature alignment assumption (\autoref{sec:preliminaries}):
diagonal entries stay positive while off-diagonal leakage is suppressed.

\section{Experimental Validation}
We validate ND-LoRA through systematic hallucination reduction experiments using parameter- and data-matched
comparisons. Although all our theoretical analysis (\autoref{sec:theory}) is model-agnostic, all empirical
results in the following sections are on a single small language model (Qwen2.5-0.5B) trained on The Pile
\citep{gao2020pile} and evaluated on 12 tasks (\autoref{tab:eval_tasks}). We describe our full experimental
setup in \autoref{sec:experimental_setup}.

\subsection{Key Results}
\begin{table}[t]
  \begin{center}
    \begin{tabular}{lcccccc}
      \multicolumn{1}{c}{\bf Model} &
      \multicolumn{1}{c}{\bf HaluEval} &
      \multicolumn{1}{c}{\bf MemoTrap} &
      \multicolumn{1}{c}{\bf TruthfulQA} &
      \multicolumn{1}{c}{\bf NQ} &
      \multicolumn{1}{c}{\bf Wikitext} &
      \multicolumn{1}{c}{\bf WG}
      \\ \hline \\[-0.75em]
      ND-LoRA R16 (P=2)        & \textbf{0.481*} & \textbf{0.666*} & \textbf{0.442*} & 0.055 & 0.784 &
      \textbf{0.574} \\
      ParScale R32 (P=2)       & 0.439 & 0.638 & 0.412 & 0.059 & 0.793 & 0.564 \\
      Qwen LoRA R32            & 0.400 & 0.634 & 0.403 & \textbf{0.065} & \textbf{0.778} & 0.572 \\
    \end{tabular}
  \end{center}
  \caption{
    \textbf{Even at $P=2$ streams, ND-LoRA achieves up to 20.2\% relative hallucination reduction vs.
    parameter-matched baseline.} Across hallucination benchmarks, ND-LoRA shows statistically
    significant improvements (HaluEval-Summarization, MemoTrap, TruthfulQA-MC2) while maintaining competitive
    Winogrande, NQ, and Wikitext BPB (lower is better) general-purpose capabilities. Baselines use higher
    LoRA ranks for parameter parity. * indicates $p < 0.05$.
  }
  \label{tab:relacc}
\end{table}

\autoref{tab:relacc} demonstrates ND-LoRA achieves substantial improvements on hallucination-sensitive
benchmarks while maintaining competitive general performance. ND-LoRA with $P=2$ streams achieves statistically
significant improvements on HaluEval-Summarization (0.481* vs 0.400, $p < 0.001$, 8.1\% absolute / 20.2\% relative),
TruthfulQA-MC2 (0.442* vs 0.403, $p = 0.030$, 3.9\% absolute / 9.5\% relative) and MemoTrap (0.666* vs 0.634, $p
< 0.001$, 3.2\% absolute / 5.1\% relative) vs parameter-matched Qwen, validating our theoretical prediction.

Although ND-LoRA's improvements specifically target reliability benchmarks, they preserve general
capabilities. Qwen slightly outperforms on Wikitext (0.778 vs. 0.784) and Natural Questions
(0.065 vs. 0.055), but ND-LoRA wins slightly on Winogrande (0.574 vs. 0.572).

Parameter efficiency is evident comparing ND-LoRA R16 ($P=2$) against Qwen2.5-0.5B LoRA R32. Despite lower-rank
adapters, ND-LoRA consistently outperforms the high-rank baseline on hallucination tasks, demonstrating that
architectural diversity provides more value at equal capacity. This shows representational diversity, not
parameter count, drives reliability gains in our experiment.

\begin{table}[t]
  \centering
  \begin{tabular}{l|l|rr}
    \bf Method & \bf Type & \bf Halluc.\ $\Delta$\% & \bf Knowledge $\Delta$\% \\
    \hline
    ND-LoRA & integrated & \textbf{+14.6\%} & +0.2\% \\
    CAD & inference-time & +4.1\% & \textbf{+1.2\%} \\
    ActDec & inference-time & +1.5\% & -2.6\% \\
    Disagreement & training-time & +1.7\% & -1.1\% \\
  \end{tabular}
  \caption{
    \textbf{ND-LoRA dominates on hallucination without a knowledge tax.} Average relative
    $\Delta$\% vs.\ the $P=1$ baseline across six hallucination (HaluEval Dial/QA/Summ, MemoTrap,
    TF-MC1/MC2) and five knowledge (NQ, PopQA, TriviaQA, Winogrande, Wikitext BPB) benchmarks.
  }
  \label{tab:baselines_summary}
\end{table}

\begin{figure}[t]
  \begin{center}
    \includegraphics[width=0.85\textwidth]{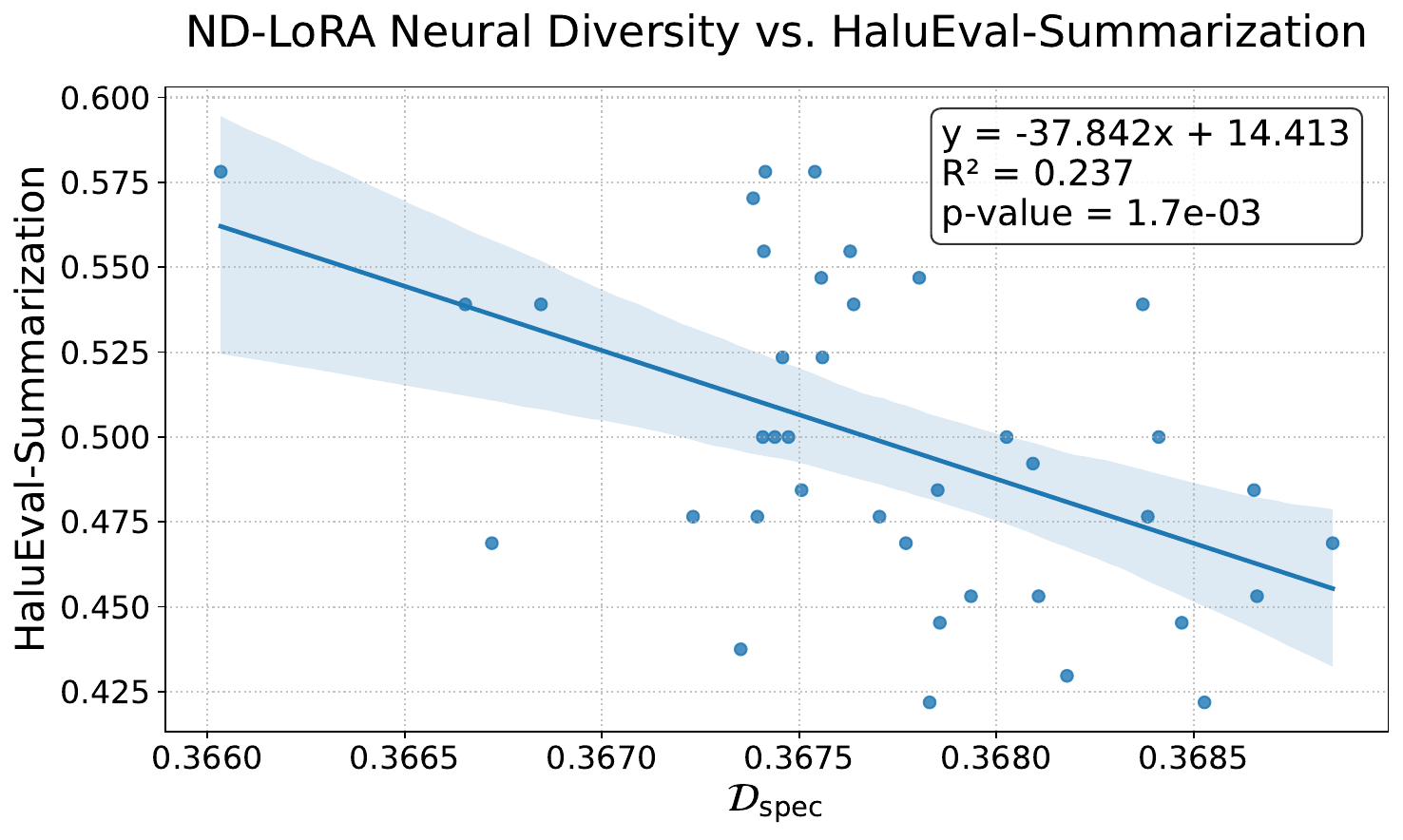}
  \end{center}
  \caption{
    \textbf{Reliability improves as neural diversity increases (lower $\mathcal{D}$).} Specifically,
    diversity ($\mathcal{D}$) is negatively correlated with HaluEval-Summarization
    performance (slope=-37.842, R²=0.237, p=0.002), consistent with $\Prob(\text{H}) \propto
    \mathcal{D}$ in \autoref{thm:halluc-with-diversity}.
  }
  \label{fig:correlational}
\end{figure}

These findings establish neural diversity as a practical reliability mechanism. Consistent improvements
across hallucination benchmarks with preserved general performance suggest ND-LoRA addresses fundamental
reliability challenges rather than metric-specific optimization. \autoref{fig:correlational} demonstrates
strong empirical correlation between neural diversity and performance, building intuition for the causal
relationship established in \autoref{sec:causality}.

\subsection{Task-Dependent Optimality}
\label{sec:taskdiv}
Further, the optimal diversity is task-dependent. \autoref{tab:taskdiv} reveals striking task-dependent
sensitivity patterns relative to the $P=1$ baseline. Hallucination-focused tasks show the largest
gains: HaluEval Summarization achieves +25.6\% relative improvement at $P=4$, HaluEval QA shows +23.4\% at $P=4$,
and TruthfulQA MC2 shows +9.5\% at $P=2$ while MemoTrap benefits from higher diversity ($P=8$, +8.8\%).
Notably, knowledge-intensive tasks like PopQA, TriviaQA and NQ show no improvement over baseline, which is
expected as ND-LoRA does not add new sources of knowledge or try to improve recall of existing knowledge.
This heterogeneity demonstrates that different tasks require different amounts of neural diversity to
maximize reliability, with hallucination-focused tasks generally benefiting most from decorrelated representations.

Disaggregating further into \emph{faithfulness} tasks (HaluEval-Dialog, -QA, -Summarization, MemoTrap~v2) and
\emph{factuality} tasks (TruthfulQA-MC1, -MC2) reveals a consistent $\approx2\times$ advantage for
faithfulness (\autoref{sec:faith-fact}). This is theoretically expected: neural diversity decorrelates the
internal verification of context-grounded claims, directly benefiting faithfulness. In contrast, factuality
requires knowledge the model may lack, which diversity alone cannot supply, limiting diversity-based gains.

\subsection{Comparison with Other Methods}
\label{sec:baseline_comparison}
\autoref{tab:baselines_summary} compares ND-LoRA to three baselines on Qwen2.5-0.5B: Context-Aware
Decoding \citep[CAD;][]{shi2024cad} and Activation Decoding \citep[ActDec;][]{chen2024actdec} are
inference-time interventions; Disagreement Regularization \citep{li2018disagreement} is a compute-matched
training-time method. ND-LoRA yields $+14.6\%$ hallucination improvement --- $3.5\times$ the next-best
baseline (CAD, $+4.1\%$) --- while keeping knowledge within $0.2\%$ of the $P=1$ baseline
(vs.\ CAD $+1.2\%$, ActDec $-2.6\%$, Disagreement $-1.1\%$). Per-benchmark scores
(\autoref{sec:baseline_detail}) show ND-LoRA winning every hallucination column; CAD edges out PopQA and
TriviaQA by $1$--$5$ basis points.

\section{Mechanistic Analysis}
\begin{table}[t]
  \centering
  \begin{tabular}{l|ccccccc}
    \textbf{Task} & \textbf{$\Delta\mathcal{D}$} & \textbf{$\Delta$ Score} & \textbf{SE} &
    \textbf{d} & \textbf{p-value} & \textbf{Sig.} & \textbf{N} \\
    \hline
    HaluEval-Summ & 0.024 & -0.005 & 0.010 & 0.007 & $1.6 \times 10^{-5}$ & *** & 512 \\
    MemoTrap v2 & 0.031 & -0.003 & 0.010 & 0.000 & $8.2 \times 10^{-5}$ & *** & 512 \\
    TruthfulQA-MC2 & 0.025 & -0.007 & 0.009 & 0.018 & $3.3 \times 10^{-7}$ & *** & 512 \\
  \end{tabular}
  \caption{
    \textbf{Artificial corruption of neural diversity establishes statistical causality.} Perturbing neural diversity
    ($\Delta\mathcal{D} > 0$) causes accuracy drops across tasks with high statistical
    significance ($p < 0.001$) via paired t-tests with Fisher meta-analysis (N=4 sub-experiments × 128 samples each).
  }
  \label{tab:causality}
\end{table}

\subsection{Neural Diversity as the Causal Mediator}
\label{sec:causality}
To establish causality beyond correlation, we perform artificial corruption interventions that directly
manipulate cross-stream similarity.

\textbf{Experiment Design.}
Starting with a pre-trained ND-LoRA $P=4$ model, we inject a corruption hook at the RMSNorm layer that
randomly substitutes the hidden state at randomly-chosen positions in a given stream from another stream, perturbing
$\mathcal{D}$ while preserving activation magnitudes. We evaluate on a matched basis: each
corrupted evaluation is paired with an uncorrupted baseline using identical samples and resampling indices.
Across 4 sub-experiments with different random seeds, we collect $N=128$ paired samples per task. This paired
design maximizes statistical power by controlling sample-level variance, analyzed via paired t-tests with
Fisher meta-analysis.

\textbf{Results.}
\autoref{tab:causality} provides statistically robust evidence that neural diversity causally affects
performance. All three tasks show highly significant accuracy drops ($p < 0.001$) when stream-level
substitution perturbs diversity ($\Delta\mathcal{D} \approx 0.025$). While effect sizes are
modest (0.3\% to 0.7\% score reduction) --- likely because artificial stream substitution creates
out-of-distribution corruption patterns --- the statistical significance establishes causality beyond
correlational association.

\subsection{Ablations}
\label{sec:ablations}
To isolate the contributions of ND-LoRA, we systematically ablate ND-LoRA components at fixed $P=4$ streams.
All variants maintain parameter parity through LoRA rank adjustments, enabling fair comparison. We measure
inference-time diversity ($\mathcal{D}$) at the aggregation layer using evaluation samples,
quantifying actual cross-stream correlation during inference.

\autoref{tab:ablations} reveals a super-linear combination: independent LoRA (+2.9\%) and Barlow Twins
(+1.4\%) sum to 4.3\% but achieve 4.9\% when combined (Stream LoRA-BT) --- a 14\% bonus. Targeting KVQ
attention amplifies this further by 2.6$\times$ to +12.8\% (ND-LoRA at fixed $P=4$; maximum gains reach
14.6\% when optimizing $P$ per-task, see \autoref{tab:taskdiv}). Neither component alone
suffices: ParScale's near-complete collapse ($\mathcal{D} = 0.9990$) yields only +0.5\%, while
Stream LoRA without regularization achieves +2.9\%, both less than a quarter of ND-LoRA's final impact. This
establishes that both architectural capacity and explicit regularization are necessary for full impact.

Notably, ParScale's original work found prefix tuning superior to LoRA for mean loss (Table~6 in
\citealt{chen2025parscale}). However, stream-aware LoRA is necessary for reducing tail probability:
even with Barlow Twins, prefix tuning collapses streams ($\mathcal{D} = 0.9988$),
while stream-aware LoRA enables decorrelation ($\mathcal{D} = 0.1530$). This illustrates
how second-moment objectives require different architectural choices than first-moment objectives.

Counterintuitively, ND-LoRA achieves best performance (+12.8\%) with \emph{higher} $\mathcal{D}$
= 0.4112 than Stream LoRA-BT's 0.1530. This reveals that strategic localization to representational
bottlenecks matters more than maximizing global decorrelation: focusing LoRA and Barlow Twins on KVQ
attention modules provides 2.6$\times$ amplification. This further reinforces how second-moment objectives
differ architecturally from first-moment ones and, consistent with \autoref{tab:taskdiv}, that neural
diversity is a task-dependent resource requiring strategic allocation to critical computational pathways.

\subsection{Practical Approximability}
While task-optimal $P_\star$ varies (\autoref{tab:taskdiv}), practitioners need not search exhaustively.
Defaulting to $P=4$ achieves 97\% of oracle hallucination performance (96\% across all 12 evaluations).
Additionally, a simple router (\autoref{sec:router_model}) achieves 99\% of oracle hallucination performance
by predicting $P$ from prompt statistics, revealing a retrieval-vs-verifiability tradeoff: question-dense
prompts favor low $P$, while longer, context-heavy prompts favor higher $P$.

\subsection{Hyperparameter Sensitivity}
Sensitivity analyses across 70+ configurations show that hallucination improvements are stable
across design layers $\ell_\star \in [7,23]$ while $\lambda_\text{BT} \in [0.01, 0.50]$ exposes a
hallucination--perplexity tradeoff (\autoref{sec:reg_sensitivity}); LoRA rank $R16$--$R128$ and alpha scalings are not
confounds and attention-only LoRA outperforms MLP-only LoRA (\autoref{sec:lora_ablations}); and even
worst-choice $P \in \{2,4,8\}$ improves hallucination over the parameter-matched baseline (\autoref{sec:sensitivity}).

\subsection{Computational Considerations}
Unlike $P$-model ensembles with $P\times$ pretraining cost, ND-LoRA achieves substantial reliability gains
at negligible overhead (1.00008$\times$ pretraining, 1.1$\times$ latency) given its single architecture. Parallelized
20M amortizes to $\approx$0.008\% of 1T-token pretraining, frozen backbone makes gradients nearly free, and
ND-LoRA requires near-identical FLOPs to ParScale at inference, with per-stream adapters servable through a
batched adapter kernel \citep{chen2023punica}. See \autoref{sec:cost_analysis}.

\begin{table}[t]
  \small
  \begin{center}
    \begin{tabular}{l|cccc|ccccc}
      \multicolumn{1}{c}{\bf Variant} &
      \multicolumn{1}{c}{\bf Streams} &
      \multicolumn{1}{c}{\bf LoRA} &
      \multicolumn{1}{c}{\bf Regul.} &
      \multicolumn{1}{c}{\bf Target} &
      \multicolumn{1}{c}{\bf $\mathcal{D}$} &
      \multicolumn{1}{c}{\bf $\overline{\Delta}$\% Score} &
      \multicolumn{1}{c}{\bf $\Delta$ Cost}
      \\ \hline \\[-0.75em]
      Standard       & 1   & Single & D      & All & --              & 0.0\%  & ~~~~~\textbf{1.0x} / \textbf{1.0x} \\
      ParScale       & $P$ & Single & D      & All & 0.9990          & +0.5\%           & 1.00008x / 1.1x \\
      ParScale-BT    & $P$ & Single & D + BT & All & 0.9988          & +1.4\%           & 1.00008x / 1.1x \\
      Stream LoRA    & $P$ & Stream & D      & All & 0.3544          & +2.9\%           & 1.00008x / 1.1x \\
      Stream LoRA-BT & $P$ & Stream & D + BT & All & \textbf{0.1530} & +4.9\%           & 1.00008x / 1.1x \\
      ND-LoRA        & $P$ & Stream & D + BT & KVQ & 0.4112          & \textbf{+12.8\%} & 1.00008x / 1.1x \\
    \end{tabular}
  \end{center}
  \caption{
    \textbf{Ablations reveal super-linear combination of impact.} Stream LoRA (+2.9\%) and Barlow
    Twins (+1.4\%) combine super-linearly (+4.9\%), and focusing on KVQ attention amplifies to +12.8\%.
    {\em LoRA}: single shared vs. $P$ stream-aware adapters.
    {\em Regularization}: Dropout vs. Barlow Twins.
    {\em Target}: All layers vs. KVQ attention only.
    {\em $\mathcal{D}$}: Neural Diversity Index (lower is better).
    {\em $\overline{\Delta}$\% Score}: avg. change (hallucination benchmarks).
    Ablations shown at fixed $P=4$ streams.
  }
  \label{tab:ablations}
\end{table}

\section{Related Work}
\textbf{Hallucination in Language Models.}
Hallucinations represent a fundamental challenge in modern language models. Comprehensive surveys establish
taxonomies that distinguish factuality vs. faithfulness \citep{huang2024survey, tonmoy2024comprehensive}.
Theoretical work proves hallucinations are mathematically inevitable in computable models under certain
resource constraints \citep{xu2024inevitable, kalai2024calibrated}, with smaller models exhibiting particular
severity on factual benchmarks \citep{lin2021truthfulqa, li2023halueval}. Mechanistic investigations reveal
hallucinations arise from internal representation failures \citep{yu2024mechanistic}, knowledge awareness limitations
\citep{ferrando2025knowledge}, and attention pattern anomalies.

Mitigation has predominantly targeted average performance.
Retrieval augmentation (RAG) incorporates external knowledge for factual grounding
\citep{niu2024ragtruth}. RLHF improves alignment \citep{bai2022constitutional}, while constitutional AI
enhances safety. Decoding methods use contrastive decoding
\citep{li2023contrastive} and classifier-free guidance \citep{sanchez2023cfg}. Critically, improving
$\E[\text{error}]$ does not guarantee improvements to $\Prob(\text{hallucination})$, as tail events
depend on variance and correlation structure, not just central tendency.

Second-moment approaches exist but lack theoretical grounding: self-consistency reduces hallucinations through
diverse sampling \citep{wang2022selfconsistency} without formal tail-probability guarantees, while deep
ensembles provide uncertainty estimates \citep{lakshminarayanan2017deep} but not hallucination-specific
bounds. We provide the first formal tail bounds connecting neural diversity to hallucination probability as a
second-moment problem.

\textbf{Deep Ensembles, Parallel Architectures \& Inference-Time Scaling}
Deep ensembles provide uncertainty estimates \citep{lakshminarayanan2017deep} with power-law scaling
\citep{lobacheva2020power} for calibration and OOD detection. LLM ensembles
benefit from explicit diversity optimization \citep{tekin2024llmtopla}, while negative
correlation learning demonstrates diversity must be actively encouraged \citep{liu1999ensemble}. The ``memory
split advantage'' shows ensembles of smaller models can outperform single large models at fixed parameter
budgets. Optimal size theory reveals weighted voting exhibits diminishing returns due to correlation and
overfitting \citep{bonab2019less}, with predictions stabilizing at 5--10 models \citep{hernandez2013ensemble}.
These approaches require multiple independent models, incurring $P\times$ training and inference costs.

Inference-time methods reduce hallucinations by reshaping the decoding distribution or diversifying over
sampling noise. Context-Aware Decoding \citep[CAD;][]{shi2024cad} and Activation Decoding
\citep[ActDec;][]{chen2024actdec} steer logits toward context-grounded generations; we benchmark ND-LoRA
head-to-head against both in \autoref{sec:baseline_comparison}. Self-consistency
\citep{wang2022selfconsistency}, confidence-based weighting \citep{taubenfeld2025confidence}, and
contrastive decoding \citep{li2023contrastive} instead aggregate over multiple generations from a fixed
representation. All of these intervene only at inference --- leaving the underlying representations unchanged ---
whereas our training-time parallelism learns coordinated streams; because the two attack at different stages
of the pipeline (representation learning vs. decoding), they compose rather than substitute.

Self-ensembled parallel architectures like ParScale \citep{chen2025parscale} break the multiplicative memory
requirements of classical ensembles by using $P$ perturbed computational pathways within a single model.
ParScale achieves $O(\log P)$ general capability gains, modeling parallel streams with correlation $\rho$ in
scaling laws $L \propto (N \cdot P^{1/\alpha} \cdot [(P-1)\rho+1]^{-1/\alpha})^{-\alpha}$. This targets mean
loss for accuracy improvements, not hallucination probability. We directly build our demonstration on
ParScale, extending their theoretical framework and implementation to tail-bound hallucinations.

\textbf{Theoretical Foundations.}
Modern portfolio theory \citep{markowitz1952portfolio} provides the mathematical
foundation for understanding correlation-based risk reduction, with diversification principles
\citep{meucci2009risk} for ensemble variance analysis. Classical ensemble theory reduces mean error
$\E[\text{loss}]$ via variance decomposition \citep{dietterich2000ensemble}.
PAC-Bayesian bounds connect diversity to minimax-optimal generalization \citep{ortega2022diversity} and
concentration inequalities showing correlation reduction tightens tail bounds
\citep{alquier2024userfriendly}. We link these frameworks to modern neural networks to bound hallucination
tail probabilities.

\textbf{Redundancy Reduction.}
A rich history of diversification exists in self-supervised learning to avoid training collapse and in PEFT
methods for efficient specialization. Self-supervised approaches like Barlow Twins \citep{zbontar2021barlow}
and VICReg \citep{bardes2022vicreg} use decorrelation to prevent dimensional collapse \citep{jing2022dimensional}.
PEFT methods like LoRA \citep{hu2022lora} and prefix-tuning \citep{li2021prefix} enable model specialization
under limited parameter budgets, with BatchEnsemble and LoRA-Ensemble achieving diversity through
parameterization \citep{wen2020batchensemble, muhlematter2024loraensemble}. We adapt these methods for
second-moment reliability guarantees.

\section{Discussion}
\label{sec:discussion}
At a time when the reliability of language models is becoming the critical barrier to real-world deployment, we
(i) provide the first formal framework to tail-bound hallucinations in ensembled language models,
demonstrating that neural diversity plays a critical role in reducing hallucinations; and, (ii) using this
technique, achieve up to 25.6\% (and 14.6\% on average) reduction in hallucination rates on evaluated benchmarks
at fixed parameter and data budgets at +0.008\% pretraining cost. Neural diversity enables reliability gains
without massive
compute scaling; the only recurring cost is a 1.1$\times$ inference latency from evaluating $P$ parallel
streams (\autoref{sec:cost_analysis}), a modest tradeoff for the reliability gains in latency-tolerant,
safety-critical deployments.

By reframing hallucinations as a second-moment problem --- controlled through variance and correlation rather
than mean optimization --- we open an under-explored research direction orthogonal to existing approaches. While
RLHF and RAG target first-moment improvements (average performance), neural diversity targets tail probability
through explicit decorrelation. This bridges portfolio theory to neural reliability, a connection
previously unexplored. The gap between extensive first-moment research and nascent second-moment
approaches (self-consistency, our work) suggests substantial opportunity for reliability-focused methods
grounded in tail-probability theory.

Our small-scale demonstration and mechanistic analysis validates the theoretical framework; scaling to
larger models is a natural next step given that continued training requires only +0.008\% additional overhead and
$P=4$ captures 94.3\% of oracle performance. The task-dependent optimal $P_\star$ in \autoref{tab:taskdiv}
reveals intriguing structure, suggesting deeper connections between task complexity, knowledge recall vs.
precision and neural diversity worthy of theoretical characterization.

Although the task-dependent optimal $P$ introduces additional complexity via a new hyperparameter, three
observations mitigate this concern. First, even at worst-case $P$, ND-LoRA still beats the parameter-matched
baseline on every hallucination benchmark (\autoref{sec:sensitivity}). Second, defaulting to $P{=}4$ captures
97\% of oracle hallucination performance with no tuning. Third, a simple two-feature router achieves 99\% of
oracle hallucination performance (\autoref{sec:router_model}), with learned coefficients reinforcing
knowledge recall vs.
precision dynamics above (\autoref{sec:faith-fact}).

Our work opens three immediate research directions: (i) \emph{Theoretical}: characterizing optimal $P_\star$
as a function of task properties via information-theoretic approaches suggested by the U-shape theorem
(\autoref{thm:u-shape}). (ii) \emph{Practical}: combining neural diversity with inference-time scaling
\citep{snell2024scaling} for multiplicative reliability gains. (iii) \emph{Extension}: preliminary work
suggests portfolio-theoretic dynamics emerge in single-stream language models via internal parallel
structures (e.g. multi-head attention), but co-exist with interference dynamics requiring a distinct
theoretical treatment. Second-moment reliability is an essential frontier as language models
become critical infrastructure in high-stakes domains.

\makeatletter
\if@accepted
\subsubsection*{Author Contributions}
Kushal Chakrabarti led the overall project design, core implementation of ND-LoRA, experimental design, and
paper writing. Nirmal Balachundhar contributed theoretical insights and implementation for novel neural
diversity techniques, experiment implementation and analysis, and paper writing.

\subsubsection*{Acknowledgments}
We thank the research community at South Park Commons for valuable discussions and feedback throughout this
project. We are grateful to Augustine Mavor-Parker, Bhav Ashok, Daniel Morillo, David Sontag, Iman Modarressi,
Jaclyn Lunger, Javier Ferrando, John Bohannon, Nelson Ray, Patrick Bozeman and Suhith Rajesh for their
insightful comments on earlier drafts. This work was supported by South Park Commons and Modal, who
generously provided compute resources for our experiments.
\fi
\makeatother

\bibliography{main}
\bibliographystyle{tmlr}

\appendix
\section{Appendix}
\subsection{Full Proofs}

\subsubsection{Proof of \autoref{lem:cosine-avg-corr}}
\begin{proof}
  We proceed in two steps. First, we show that the shared linear readout $A$ can at most distort cosines
  between representations by a factor of $\kappa^2$. Second, we convert bounded cosine
  alignment between the error vectors $\xi_i$ into a bound on average noise correlation.

  \paragraph{Step 1: local linear readout and cosine distortion.}
  Let $s_{\min}$ and $s_{\max}$ denote the minimal and maximal singular values of $A$ and
  $\kappa = s_{\max}/s_{\min}$ its condition number.
  For streams $i,j$ with representations $\tilde{z}_i, \tilde{z}_j \in \mathbb{R}^d$, the local linear
  readout gives $\xi_i = A\tilde{z}_i$ and $\xi_j = A\tilde{z}_j$.
  Then
  \[
    \cos\angle(\xi_i,\xi_j)
    \;=\;
    \frac{\langle \xi_i,\xi_j\rangle}{\|\xi_i\|_2 \,\|\xi_j\|_2}
    \;=\;
    \frac{\tilde{z}_i^\top A^\top A\, \tilde{z}_j}{\|A \tilde{z}_i\|_2 \,\|A \tilde{z}_j\|_2}.
  \]
  Write $\tilde{z}_i^\top A^\top A\, \tilde{z}_j = \sum_k s_k^2 a_k b_k$ where $a_k =
  e_k^\top \tilde{z}_i$, $b_k = e_k^\top \tilde{z}_j$, $s_k$ are the singular values of $A$, and $\{e_k\}$
  are its right singular vectors (equivalently, eigenvectors of $A^\top A$ with eigenvalues $s_k^2$).
  By feature alignment (\autoref{sec:preliminaries}), $a_k b_k \ge 0$ for all $k$, so:
  \[
    |\tilde{z}_i^\top A^\top A\, \tilde{z}_j|
    \;=\;
    \sum_k s_k^2 a_k b_k
    \;\le\;
    s_{\max}^2 \sum_k a_k b_k
    \;=\;
    s_{\max}^2\,\tilde{z}_i^\top \tilde{z}_j.
  \]
  When $\tilde{z}_i \perp \tilde{z}_j$, both sides are zero, resolving the orthogonal case cleanly.
  For the denominator,
  \[
    \|A \tilde{z}_i\|_2 \,\|A \tilde{z}_j\|_2
    \;\ge\;
    s_{\min}^2\,\|\tilde{z}_i\|_2 \,\|\tilde{z}_j\|_2.
  \]
  Combining,
  \[
    |\cos\angle(\xi_i,\xi_j)|
    \;\le\;
    \frac{s_{\max}^2}{s_{\min}^2}
    \frac{|\tilde{z}_i^\top \tilde{z}_j|}{\|\tilde{z}_i\|_2 \,\|\tilde{z}_j\|_2}
    \;=\;
    \kappa^2\,|\cos\angle(\tilde{z}_i,\tilde{z}_j)|.
  \]
  Squaring both sides yields, for every pair of streams $(i,j)$ and every input $x$ with nonzero norms,
  \begin{equation}
    \label{eq:cosine-distortion}
    \cos^2\angle\big(\xi_i(x),\xi_j(x)\big)
    \;\le\;
    \kappa^4\,
    \cos^2\angle\big(\tilde{z}_i(x),\tilde{z}_j(x)\big).
  \end{equation}
  Taking expectations over $x$ gives
  \[
    \mathcal{D}_{\xi,ij}^2
    \;\triangleq\;
    \E_x\big[\cos^2\angle(\xi_i(x),\xi_j(x))\big]
    \;\le\;
    \kappa^4\,
    \E_x\big[\cos^2\angle(\tilde{z}_i(x),\tilde{z}_j(x))\big]
    \;\triangleq\;
    \kappa^4\,\mathcal{D}_{ij}^2,
  \]
  where $\mathcal{D}_{ij}^2$ denotes the pairwise cosine diversity in representation space.
  Averaging over pairs and taking square roots yields
  \begin{equation}
    \label{eq:Dxi-vs-Dz}
    \mathcal{D}_\xi
    \;\triangleq\;
    \sqrt{\E_{i<j} \mathcal{D}_{\xi,ij}^2}
    \;\le\;
    \kappa^2\,
    \sqrt{\E_{i<j} \mathcal{D}_{ij}^2}
    \;=\;
    \kappa^2\,\mathcal{D}.
  \end{equation}

  \paragraph{Step 2: from cosine alignment to average correlation.}
  We now bound the average correlation $\bar{\rho}$ in terms of $\mathcal{D}_\xi$.
  Fix a pair $(i,j)$ and write
  \[
    X \;\triangleq\; \langle \xi_i,\xi_j\rangle,
    \qquad
    B \;\triangleq\; \|\xi_i\|_2\,\|\xi_j\|_2.
  \]
  Whenever $B>0$,
  \[
    \cos\angle(\xi_i,\xi_j)
    \;=\;
    \frac{X}{B},
    \qquad
    \mathcal{D}_{\xi,ij}^2
    \;=\;
    \E\!\left[\left(\frac{X}{B}\right)^2\right].
  \]
  Using Cauchy--Schwarz with $U = X/B$ and $V = B$, we obtain
  \[
    \Sigma_{ij}^2
    =
    \big(\E[X]\big)^2
    =
    \big(\E[UV]\big)^2
    \;\le\;
    \E[U^2]\,\E[V^2]
    =
    \mathcal{D}_{\xi,ij}^2\,\E\big[\|\xi_i\|_2^2\,\|\xi_j\|_2^2\big].
  \]
  Apply Cauchy--Schwarz again to the norms and use the kurtosis bound:
  \[
    \E\big[\|\xi_i\|_2^2\,\|\xi_j\|_2^2\big]
    \;\le\;
    \sqrt{\E[\|\xi_i\|_2^4]\,\E[\|\xi_j\|_2^4]}
    \;\le\;
    \sqrt{C_4\sigma_i^4\,C_4\sigma_j^4}
    =
    C_4\,\sigma_i^2\sigma_j^2.
  \]
  Combining,
  \[
    \Sigma_{ij}^2
    \;\le\;
    C_4\,\mathcal{D}_{\xi,ij}^2\,\sigma_i^2\sigma_j^2,
    \qquad
    \rho_{ij}^2
    =
    \frac{\Sigma_{ij}^2}{\sigma_i^2\sigma_j^2}
    \;\le\;
    C_4\,\mathcal{D}_{\xi,ij}^2,
  \]
  so
  \begin{equation}
    \label{eq:rhoij-vs-Dxi}
    |\rho_{ij}|
    \;\le\;
    \sqrt{C_4}\,\mathcal{D}_{\xi,ij}.
  \end{equation}

  Finally, average over pairs and apply Cauchy--Schwarz in the index space:
  \[
    |\bar{\rho}|
    =
    \left|\E_{i<j}[\rho_{ij}]\right|
    \;\le\;
    \E_{i<j}[|\rho_{ij}|]
    \;\le\;
    \sqrt{C_4}\,\E_{i<j}[\mathcal{D}_{\xi,ij}]
    \;\le\;
    \sqrt{C_4}\,\sqrt{\E_{i<j} \mathcal{D}_{\xi,ij}^2}
    =
    \sqrt{C_4}\,\mathcal{D}_\xi.
  \]
  Plugging \eqref{eq:Dxi-vs-Dz} into this inequality gives
  \[
    |\bar{\rho}|
    \;\le\;
    \sqrt{C_4}\,\mathcal{D}_\xi
    \;\le\;
    \sqrt{C_4}\,\kappa^2\,\mathcal{D}
    \;=\;
    C_\ast\,\mathcal{D},
  \]
  with $C_\ast = \sqrt{C_4}\,\kappa^2$, as claimed.
\end{proof}

\subsubsection{Proof of \autoref{thm:halluc-with-diversity}}
\begin{proof}
  We work under the signal--noise model from the preliminaries. By \citealt{markowitz1952portfolio},
  \[
    \Var\left(\frac{1}{P} \sum_{i=0}^{P-1} X_i\right) = \bar{\sigma}^2\left(\frac{1-\rho}{P} + \rho\right)
  \]
  where $\bar{\sigma}^2 = \frac{1}{P}\sum_i \Var(X_i)$ is the average variance and $\rho = \E_{i \neq
  j}[\Corr(X_i, X_j)]$ is the average pairwise correlation.

  From \autoref{lem:cosine-avg-corr}, we have
  \[
    |\bar{\rho}| \;\le\; C_\ast\,\mathcal{D}.
  \]

  Substituting this into the variance expression yields
  \begin{align*}
    \Var(E_w)
    &\le \bar{\sigma}^2\left(
      \frac{1 - C_\ast\,\mathcal{D}}{P}
      + C_\ast\,\mathcal{D}
    \right),
  \end{align*}
  which is exactly the claimed variance bound in \autoref{thm:halluc-with-diversity}.

  By construction, the hallucination event is
  \[
    \HALL \;\triangleq\; \{E_w \ge \delta\},
    \qquad \delta > 0,
  \]
  and we have already noted that $\E[E_w]=0$.
  Applying the one-sided Chebyshev inequality from the preliminaries
  to the random variable $E_w$ with mean $0$ and variance
  $v = \Var(E_w)$ gives
  \[
    \Prob(\HALL)
    \;=\;
    \Prob(E_w \ge \delta)
    \;\le\;
    \frac{v}{v + \delta^2}
    \;=\;
    \frac{\Var(E_w)}
    {\Var(E_w) + \delta^2}.
  \]

  Substituting $\Var(E_w)$ by its upper bound yields
  \[
    \Prob(\HALL)
    \;\le\;
    \frac{\bar{\sigma}^2 \left(
        \frac{1 - C_\ast\,\mathcal{D}}{P}
        +
        C_\ast\,\mathcal{D}
    \right)}
    {\bar{\sigma}^2 \left(
        \frac{1 - C_\ast\,\mathcal{D}}{P}
        +
        C_\ast\,\mathcal{D}
      \right)
    + \delta^2}.
  \]
  Dividing both numerator and denominator by $\bar{\sigma}^2$ matches the bound stated in
  \autoref{thm:halluc-with-diversity}.
\end{proof}

\subsubsection{Proof of \autoref{thm:u-shape}}
\begin{proof}
  Extend $P$ to a real variable with domain $P\ge 1$; the claim for integer $P$ follows by restriction.

  Under uniform weights $w_i = 1/P$, the ensemble error variance can be written as
  \[
    v(P)
    \triangleq
    \Var(E_w)
    =
    \bar{\sigma}^2\left(\frac{1-\bar{\rho}(P)}{P} + \bar{\rho}(P)\right),
  \]
  with $\bar{\sigma}^2>0$ and
  \[
    \bar{\rho}(P)
    =
    \rho_0 + \beta (P-1)^\gamma,
    \qquad
    \rho_0\in[0,1),\;\beta>0,\;\gamma>0.
  \]
  The bound from the main text is
  \[
    \Prob(\HALL)
    \le
    \mathcal{B}(P)
    \triangleq
    \frac{v(P)}{v(P)+\delta^2},
    \qquad \delta>0.
  \]

  \paragraph{Step 1: Reduction to $v(P)$.}
  Define $\phi(x) \triangleq x/(x+\delta^2)$ for $x\ge 0$. Then
  \[
    \phi'(x)
    =
    \frac{\delta^2}{(x+\delta^2)^2}
    >0,
  \]
  so $\phi$ is strictly increasing. Hence $\mathcal{B}(P) = \phi\big(v(P)\big)$ has the same extrema and monotonicity
  as $v(P)$. Since $\bar{\sigma}^2>0$, it suffices to analyze
  \[
    f(P)
    \triangleq
    \frac{v(P)}{\bar{\sigma}^2}
    =
    \frac{1-\bar{\rho}(P)}{P} + \bar{\rho}(P).
  \]

  \paragraph{Step 2: First derivative and unique critical point.}
  For $P>1$,
  \[
    \bar{\rho}'(P)
    =
    \beta\gamma(P-1)^{\gamma-1}.
  \]
  A direct calculation gives
  \begin{align*}
    f'(P)
    &= \frac{\mathrm{d}}{\mathrm{d}P}
    \Big(\tfrac{1-\bar{\rho}(P)}{P} + \bar{\rho}(P)\Big) \\
    &= \frac{\beta(P-1)^\gamma(P\gamma + 1) + (\rho_0 - 1)}{P^2}
    \triangleq
    \frac{N(P)}{P^2}.
  \end{align*}
  We study $N(P)$.

  At $P=1$ we have
  \[
    N(1)
    = \beta\cdot 0^\gamma(\gamma+1) + (\rho_0 - 1)
    = \rho_0 - 1 < 0.
  \]
  Differentiating $N$ for $P>1$ yields
  \[
    N'(P)
    = \beta\gamma(\gamma+1) \cdot P \cdot (P-1)^{\gamma-1}.
  \]
  All factors on the right are strictly positive for $P>1$, so $N'(P)>0$ on $(1,\infty)$ and $N$ is strictly
  increasing. Moreover,
  \[
    (P-1)^\gamma(P\gamma+1)
    \sim \gamma P^{\gamma+1}
    \xrightarrow[P\to\infty]{} \infty,
  \]
  so $N(P)\to +\infty$ as $P\to\infty$. By continuity and strict monotonicity, there exists a unique
  $P_\star>1$ such that $N(P_\star)=0$.

  Because $P^2>0$ for all $P\ge 1$, the sign of $f'(P)$ matches that of $N(P)$:
  \[
    f'(P)
    \begin{cases}
      < 0, & 1 < P < P_\star, \\[2pt]
      = 0, & P = P_\star, \\[2pt]
      > 0, & P > P_\star.
    \end{cases}
  \]
  Thus $f$ (and hence $v$) is strictly decreasing on $(1,P_\star)$ and strictly increasing on $(P_\star,\infty)$;
  $P_\star$ is the unique global minimizer.

  \paragraph{Step 3: U-shape of the hallucination bound.}
  Since $\mathcal{B}(P)=\phi\big(v(P)\big)$ and $\phi$ is strictly increasing,
  \[
    \mathcal{B}'(P)
    = \phi'(v(P)) \cdot v'(P),
    \qquad
    \phi'(v(P))>0,
  \]
  so $\mathcal{B}$ inherits the same monotonicity:
  it is strictly decreasing on $(1,P_\star)$, strictly increasing on $(P_\star,\infty)$, and
  \[
    \mathcal{B}(P_\star)
    = \min_{P\ge 1} \mathcal{B}(P).
  \]
  Therefore the upper bound on $\Prob(\HALL)$ is U-shaped in $P$ with a unique global minimum at
  $P_\star$, determined by the parameters $(\rho_0,\beta,\gamma)$ governing $\bar{\rho}(P)$.
\end{proof}

\subsection{Training Cost and Latency Analysis}
\label{sec:cost_analysis}

Our cost accounting adopts the analysis framework of the original ParScale work \citep{chen2025parscale},
which established that $P$ parallel streams add far less inference cost than parameter scaling and quantified
this with an llm-analysis-based cost model. Applying that framework to ND-LoRA, three factors keep overhead
negligible: (1) fine-tuning on 20M tokens amortizes to approximately 0.008\% of 1T-token pretraining, (2) the
frozen backbone makes backward passes nearly free, and (3) inference keeps FLOPs near-identical to ParScale
through multi-tenant LoRA serving \citep{chen2023punica}. We cross-check our inference-latency figures
directly against ParScale's own cost analyzer on our configuration (\autoref{sec:cost_analysis}, below).

\subsubsection{Cost Model}

\noindent\textbf{Standard Fine-Tuning (P=1) Baseline.}
Consider a standard LoRA fine-tuning setup with 495M backbone parameters frozen and 1.3M trainable adapter
parameters. We express each term both as a multiple of the baseline forward pass --- one cost unit
$= 2N_\text{params} \approx 9.9\times10^8$ FLOP/token for the 495M backbone --- and in absolute FLOP/token.
A typical training step consists of:
\begin{itemize}
  \item \textbf{Forward pass}: $1.0\times$ cost through 495M parameters ($\approx 9.9\times10^8$ FLOP/token)
  \item \textbf{Backward pass}: $2.0 \times (1.3\text{Y} / 495\text{Y}) \approx 0.005\times$ cost (frozen backbone:
    gradients propagate only through the 1.3M trainable adapter parameters; $\approx 5\times10^6$ FLOP/token)
  \item \textbf{Total baseline}: 1.005 cost units per training step ($\approx 1.0\times10^9$ FLOP/token)
\end{itemize}

\noindent\textbf{ND-LoRA (P=4) Fine-Tuning.}
With $P=4$ parallel streams, ND-LoRA processes data through multiple independent pathways:
\begin{itemize}
  \item \textbf{Forward pass}: $4.0\times$ cost (P parallel forward passes through full 495M model;
    $\approx 4.0\times10^9$ FLOP/token)
  \item \textbf{Backward pass}: $2.0 \times (1.3\text{Y} / 495\text{Y}) \approx 0.005\times$ cost (gradients only
    propagate through 1.3M trainable parameters after aggregation; $\approx 5\times10^6$ FLOP/token)
  \item \textbf{Barlow Twins regularization}: $0.01\times$ cost (six $d\times d$ cross-correlations at one
      design layer across P choose 2 streams and whitening: $6\times 2d^2 \approx 9.6\times10^6$ FLOP/token,
    $\approx$1\% of a single forward pass)
  \item \textbf{Prefix/aggregator overhead}: $0.05\times$ cost (additional trainable components;
    $\approx 5\times10^7$ FLOP/token)
\end{itemize}

\paragraph{Amortization over pretraining.}
As demonstrated by ParScale \citep{chen2025parscale}, our technique does not need to run over the full
pretraining duration to be effective; it's sufficient to run for a brief fine-tuning period after the core
pretraining run. Amortized over a typical 1T-token pretraining budget, the 20M fine-tuning tokens add only
$\approx$80.9M token-equivalents (\autoref{tab:cost-breakdown}), so $(1\text{T} + 80.9\text{M}) / 1\text{T}
\approx 1.00008\times$ --- approximately 0.008\% incremental overhead over the full training lifecycle.

\subsubsection{Summary of All Variants During Training}
\autoref{tab:cost-breakdown} shows the complete cost breakdown for all ablation variants.

\begin{table}[ht]
  \centering
  \small
  \begin{tabular}{l|cccc|cc}
    \textbf{Variant} & \textbf{Forward} & \textbf{Backward} & \textbf{BT} & \textbf{Other} & \textbf{Per-Step} &
    \textbf{Total} \\
    \hline
    Standard       & 1.0 & 0.005 & 0.0  & 0.0  & 1.005 & 1.00002× \\
    ParScale       & 4.0 & 0.005 & 0.0  & 0.01 & 4.015 & 1.00008× \\
    ParScale-BT    & 4.0 & 0.005 & 0.01 & 0.01 & 4.025 & 1.00008× \\
    Indep. LoRA    & 4.0 & 0.005 & 0.0  & 0.05 & 4.055 & 1.00008× \\
    ND-LoRA        & 4.0 & 0.005 & 0.01 & 0.05 & 4.065 & 1.00008× \\
  \end{tabular}
  \caption{Fine-tuning cost breakdown (20M tokens). \emph{Forward}: P parallel passes through 495M backbone.
    \emph{Backward}: gradients through the 1.3M trainable parameters only (frozen backbone). \emph{BT}: Barlow
    Twins correlation computation. \emph{Other}: prefix/aggregator overhead. \emph{Per-Step}: per-step cost in
    baseline-forward units. \emph{Total}: total lifecycle cost over 1T-token pretraining, charging the 20M
  fine-tuning tokens at each variant's \emph{Per-Step} relative to the standard baseline.}
  \label{tab:cost-breakdown}
\end{table}

\subsubsection{Summary of Inference Latency}
Running ParScale's own cost analyzer \citep{chen2025parscale} on our configuration (Qwen2.5-0.5B, 24 layers,
batch~1, 64+64 tokens) yields per-token latency relative to the $P{=}1$ backbone of $1.03\times$ ($P{=}2$),
$1.06\times$ ($P{=}4$), and $1.14\times$ ($P{=}8$); memory rises by only $2$--$4\%$. This estimate is driven
by three key properties:
\begin{itemize}
  \item \textbf{Parameter parity}: All variants maintain identical total parameter counts by adjusting LoRA rank
  \item \textbf{Parallel processing}: P streams process in parallel; latency dominated by slowest stream + aggregation
  \item \textbf{Dynamic loading}: Stream-specific LoRA adapters can be served by a batched (grouped)
    adapter kernel \citep{chen2023punica} rather than merged into $P$ weight copies, letting all $P$ adapters
    execute in a single kernel launch and ride the same batched forward
  \item \textbf{Aggregation overhead}: Lightweight MLP aggregator adds $\approx$10\% latency
\end{itemize}

These multipliers are identical across ParScale, ParScale-BT, Indep. LoRA, and ND-LoRA because inference does
not involve Barlow Twins regularization and all parameter operations are equivalent.

Our $1.1\times$ estimate comes from the ParScale analyzer under the assumption that ND-LoRA is served the
way systems already serve many LoRAs at once, with a batched adapter kernel \citep{chen2023punica} that runs
all $P$ adapters together in one pass and so adds nothing beyond the shared backbone the analyzer already
counts. Served naively, one stream at a time, latency would instead climb toward $P\times$; served using
standard techniques, ND-LoRA matches ParScale's $1.03$--$1.14\times$ latency.

\subsubsection{Summary}

\begin{itemize}
  \item \textbf{Per-step training overhead}: $\approx$4.04× per step, dominated by the $P{=}4$ parallel forward passes
    (backward nearly free for both baseline and method)
  \item \textbf{Total cost}: $\approx0.008\%$ when amortized over 1T-token pretraining
  \item \textbf{Inference latency}: 1.1× across all $P \ge 1$ variants with parameter matching
  \item \textbf{Practical impact}: Negligible computational overhead for 25.6\% hallucination reduction
\end{itemize}

\subsection{Use of Large Language Models}
Large language models were used as a compilation tool to assist with writing and organizing sections of this
paper, including literature review synthesis, section structuring, LaTeX formatting, and co-generation of
experimental code. All technical content, experimental design, theoretical contributions, and scientific
claims are the authors' original work. The models served primarily to improve clarity, organization, and
implementation of our ideas rather than generate novel scientific insights.

\subsection{Experimental Setup}
\label{sec:experimental_setup}

\textbf{Model and Architecture.} We use Qwen2.5-0.5B (896 hidden dimensions, 24 layers) with ND-LoRA
across $P \in \{1, 2, 4, 8\}$ parallel streams applied to QKV self-attention modules and a design layer of
20 for de-correlation loss. Each stream uses independent rank-16 LoRA adapters and 48 prefix tokens, totaling
5-20M trainable parameters with 495M backbone frozen. Baseline methods use higher-rank LoRA (R32-R128) for
parameter matching.

\textbf{Training Protocol.} Models train on 20M tokens from The Pile \citep{gao2020pile} (8 random shards,
fixed seeds). We use
1024-token sequences, AdamW optimization (peak lr 3e-4, cosine decay, 2\% warmup), batch size 64, bfloat16
precision. Training completes in $\approx$5K steps ($\approx$30 min. on A100).

\textbf{Evaluation Benchmarks.} We evaluate on \textbf{12 tasks} drawn from 9 source datasets
(\autoref{tab:eval_tasks}), reported separately throughout
(\autoref{tab:results_p2}--\ref{tab:results_p8}). Two datasets contribute more than one task, because they
define distinct scored splits (HaluEval) or distinct scoring protocols (TruthfulQA).

\begin{table}[htbp]
  \centering
  \begin{tabular}{lll}
    \textbf{Category} & \textbf{Dataset} & \textbf{Task} \\
    \hline
    \multirow{6}{*}{Hallucination (6)}
    & \multirow{3}{*}{HaluEval} & HE-Dialog \\
    & & HE-QA \\
    & & HE-Summ \\
    & MemoTrap~v2 & MemoTrap \\
    & \multirow{2}{*}{TruthfulQA} & TF-MC1 \\
    & & TF-MC2 \\
    \hline
    \multirow{4}{*}{Knowledge (4)}
    & Natural Questions & NQ-8 (8-shot) \\
    & NQ-swap & NQ-swap \\
    & TriviaQA & TQA-8 (8-shot) \\
    & PopQA & PopQA \\
    \hline
    \multirow{2}{*}{General (2)}
    & WikiText & Wikitext BPB \\
    & WinoGrande & Winogrande \\
  \end{tabular}
  \caption{
    \textbf{The 12 evaluation tasks, drawn from 9 source datasets.} Hallucination tasks measure the failure
    mode our theory targets; knowledge and general tasks verify that reliability gains do not come at the
    cost of factual recall or core capability.
  }
  \label{tab:eval_tasks}
\end{table}

Six \emph{hallucination} tasks measure the failure mode our theory targets. HaluEval
\citep{li2023halueval} contributes three, detecting hallucinated dialogue responses (HE-Dialog), answers
unsupported by context (HE-QA), and unfaithful summaries (HE-Summ); MemoTrap~v2
\citep{mckenzie2023inverse} tests instruction-following against a memorized continuation; and TruthfulQA
\citep{lin2021truthfulqa} poses common misconceptions, scored on the single true answer (TF-MC1) or mass
across all true answers (TF-MC2).

Four \emph{knowledge} tasks verify factual recall: Natural Questions \citep[NQ-8,
8-shot;][]{kwiatkowski2019natural} and TriviaQA \citep[TQA-8, 8-shot;][]{joshi2017triviaqa} for
open-domain recall, PopQA \citep{mallen2023trust} for long-tail entities, and NQ-swap
\citep{longpre2021entity}, an adversarial variant that substitutes entities counterfactually. Two
\emph{general} tasks guard the frozen backbone: Wikitext BPB \citep{merity2017pointer} for language modeling
quality and Winogrande \citep{sakaguchi2020winogrande} for commonsense coreference.

Together, the three groups separate the effects our claims depend on: reduction of correlated
representational error, preservation of stored knowledge, and preservation of general capability. All tasks
are run through the Hallucinations Leaderboard evaluation suite \citep{hong2024hallucinations}, built on the
EleutherAI LM Evaluation Harness \citep{gao2023evalharness}, ensuring standardized prompting and scoring.

\textbf{Neural Diversity Measurement.} We compute $\mathcal{D}$ at the final RMSNorm layer by first whitening
representations per feature dimension across batch and sequence positions (zero mean, unit variance), then
computing pairwise cosine similarity between streams. This is equivalent to the Barlow Twins
cross-correlation formulation (Eq.~2 in \citet{zbontar2021barlow}) when features are whitened.

\textbf{Statistical Methodology.} We evaluate significance using McNemar's test for binary classification
tasks and two-tailed bootstrap tests with 10,000 samples for other tasks. Improvements marked with * are
significant at $p < 0.05$.

\subsection{Complete Benchmark Results}
\label{sec:complete_results}

Tables \ref{tab:results_p2}--\ref{tab:results_p8} provide comprehensive results across $P \in \lbrace 1, 2,
4, 8 \rbrace$ configurations with parameter-matched $P=1$ baselines. This complete view demonstrates the
thoroughness of our evaluation and enables independent verification of claims in the main text.

\begin{table}[htbp]
  \centering
  \begin{tabular}{l|ccc}
    \textbf{Evaluation} & \textbf{Qwen LoRA} & \textbf{ParScale} & \textbf{ND-LoRA} \\
    \hline
    HE Dialog & 0.458 & 0.453 & \textbf{0.513} \\
    HE QA & 0.365 & 0.337 & \textbf{0.406} \\
    HE Summ & 0.400 & 0.439 & \textbf{0.481} \\
    MemoTrap & 0.634 & 0.638 & \textbf{0.666} \\
    NQ-8 & \textbf{0.065} & 0.059 & 0.055 \\
    TQA-8 & \textbf{0.188} & 0.185 & 0.160 \\
    TF-MC1 & 0.251 & 0.259 & \textbf{0.269} \\
    TF-MC2 & 0.403 & 0.412 & \textbf{0.442} \\
    NQ-swap & \textbf{0.550} & 0.546 & 0.528 \\
    PopQA & \textbf{0.111} & 0.109 & 0.101 \\
    Wikitext BPB & \textbf{0.775} & 0.797 & 0.797 \\
    Winogrande & 0.572 & 0.564 & \textbf{0.574} \\
  \end{tabular}
  \caption{
    Benchmark results for $P=2$ (Qwen R32) parameter-matched models.
  }
  \label{tab:results_p2}
\end{table}

\begin{table}[htbp]
  \centering
  \begin{tabular}{l|ccc}
    \textbf{Evaluation} & \textbf{Qwen LoRA} & \textbf{ParScale} & \textbf{ND-LoRA} \\
    \hline
    HE Dialog & 0.464 & 0.459 & \textbf{0.516} \\
    HE QA & 0.341 & 0.322 & \textbf{0.451} \\
    HE Summ & 0.394 & 0.409 & \textbf{0.502} \\
    MemoTrap & 0.629 & 0.634 & \textbf{0.635} \\
    NQ-8 & \textbf{0.065} & 0.061 & 0.059 \\
    TQA-8 & \textbf{0.191} & 0.185 & 0.172 \\
    TF-MC1 & 0.245 & 0.253 & \textbf{0.262} \\
    TF-MC2 & 0.399 & 0.413 & \textbf{0.416} \\
    NQ-swap & \textbf{0.554} & 0.542 & 0.535 \\
    PopQA & \textbf{0.110} & \textbf{0.110} & 0.106 \\
    Wikitext BPB & \textbf{0.778} & 0.793 & 0.795 \\
    Winogrande & 0.564 & 0.573 & \textbf{0.577} \\
  \end{tabular}
  \caption{
    Benchmark results for $P=4$ (Qwen R64) parameter-matched models.
  }
  \label{tab:results_p4}
\end{table}

\begin{table}[htbp]
  \centering
  \begin{tabular}{l|ccc}
    \textbf{Evaluation} & \textbf{Qwen LoRA} & \textbf{ParScale} & \textbf{ND-LoRA} \\
    \hline
    HE Dialog & 0.460 & 0.465 & \textbf{0.475} \\
    HE QA & 0.344 & 0.335 & \textbf{0.370} \\
    HE Summ & 0.379 & 0.416 & \textbf{0.450} \\
    MemoTrap & 0.630 & 0.639 & \textbf{0.689} \\
    NQ-8 & \textbf{0.066} & 0.063 & 0.059 \\
    TQA-8 & \textbf{0.192} & 0.182 & 0.171 \\
    TF-MC1 & 0.251 & 0.256 & \textbf{0.259} \\
    TF-MC2 & 0.407 & 0.414 & \textbf{0.424} \\
    NQ-swap & 0.551 & 0.540 & \textbf{0.554} \\
    PopQA & \textbf{0.110} & 0.109 & 0.103 \\
    Wikitext BPB & \textbf{0.778} & 0.779 & 0.784 \\
    Winogrande & 0.569 & \textbf{0.577} & 0.568 \\
  \end{tabular}
  \caption{
    Benchmark results for $P=8$ (Qwen R128) parameter-matched models.
  }
  \label{tab:results_p8}
\end{table}


\subsection{Inference-Time and Training-Time Baseline Details}
\label{sec:baseline_detail}

\autoref{tab:baselines_detail} reports the per-benchmark absolute scores underlying the summary in
\autoref{tab:baselines_summary}, using the same evaluation harness and Qwen2.5-0.5B backbone as the
baseline rows. CAD, ActDec, and Disagreement numbers come from an independent evaluation of the same
benchmarks \citep{shi2024cad, chen2024actdec, li2018disagreement} (3 seeds, averaged). HellaSwag and
MMLU were not evaluated in our harness and are omitted; all other tasks common to both evaluation
pipelines are reported. We read CAD/ActDec/Disagreement's $\text{nq8}$ and $\text{tqa8}$ scores directly
from their cached per-seed result files; these were computed at evaluation time but excluded from the
baselines' paper table via a task blacklist.

\begin{table}[htbp]
  \centering
  \begin{tabular}{l|ccccc}
    \textbf{Evaluation} & \textbf{Qwen LoRA R32} & \textbf{ND-LoRA} & \textbf{CAD$^\dagger$} &
    \textbf{ActDec$^\dagger$} & \textbf{Disagreement$^\ddagger$} \\
    \hline
    HE-Dial & 0.458 & \textbf{0.516} & 0.476 & 0.458 & 0.466 \\
    HE-QA & 0.365 & \textbf{0.451} & 0.411 & 0.376 & 0.388 \\
    HE-Summ & 0.400 & \textbf{0.502} & 0.494 & 0.463 & 0.467 \\
    MemoTrap & 0.634 & \textbf{0.689} & 0.640 & 0.642 & 0.651 \\
    TF-MC1 & 0.251 & \textbf{0.269} & 0.254 & 0.251 & 0.250 \\
    TF-MC2 & 0.403 & \textbf{0.442} & 0.392 & 0.416 & 0.393 \\
    NQ & 0.065 & 0.065 & 0.066 & \textbf{0.066} & 0.063 \\
    PopQA & 0.111 & 0.111 & \textbf{0.112} & 0.112 & 0.111 \\
    TriviaQA & 0.188 & 0.188 & \textbf{0.196} & 0.191 & 0.184 \\
    Wikitext & \textbf{0.775} & \textbf{0.775} & 0.779 & 0.923 & 0.776 \\
    WG & 0.572 & \textbf{0.577} & 0.569 & 0.575 & 0.566 \\
  \end{tabular}
  \caption{
    \textbf{Per-benchmark scores underlying \autoref{tab:baselines_summary}.} Inference-time
    ($\dagger$) applied to pretrained Qwen2.5-0.5B; training-time ($\ddagger$) parameter-matched
    to our backbone \citep{shi2024cad, chen2024actdec, li2018disagreement}. Wikitext is BPB (lower
    is better).
  }
  \label{tab:baselines_detail}
\end{table}

\subsection{Faithfulness vs.\ Factuality Disaggregation}
\label{sec:faith-fact}

To understand which hallucination modes neural diversity addresses, we disaggregate benchmarks into
faithfulness (HaluEval-Dialog, -QA, -Summarization, MemoTrap~v2) and factuality (TruthfulQA-MC1, -MC2) tasks.
\autoref{tab:faith-fact} shows that faithfulness gains are roughly twice as large as factuality gains across
all $P$. This matches our theoretical framing: neural diversity decorrelates internal verification of
context-grounded claims, directly benefiting faithfulness, whereas factuality gains are smaller because
diversity cannot supply knowledge the model lacks.

\begin{table}[h]
  \centering
  \begin{tabular}{l|cccc}
    \textbf{Category} & \textbf{P=2} & \textbf{P=4} & \textbf{P=8} & \textbf{Oracle $P^\star$} \\
    \hline
    Faithfulness (4 tasks) & +13.2\% & +16.2\% & +9.0\% & +17.9\% \\
    Factuality (2 tasks) & +8.4\% & +4.4\% & +4.7\% & +8.4\% \\
    Ratio & 1.6$\times$ & 3.7$\times$ & 1.9$\times$ & 2.1$\times$ \\
  \end{tabular}
  \caption{Relative hallucination reduction by category, disaggregated into faithfulness (HaluEval-Dialog,
    -QA, -Summarization, MemoTrap~v2) and factuality (TruthfulQA-MC1, -MC2) tasks, across $P \in \{2, 4, 8\}$
  and oracle $P^\star$.}
  \label{tab:faith-fact}
\end{table}

\subsection{Sensitivity to Suboptimal $P$}
\label{sec:sensitivity}

A practical concern with task-dependent optimal $P_\star$ is robustness: what happens when the deployed $P$
does not match the task-optimal value? \autoref{tab:sensitivity} quantifies this by measuring, for each task,
the performance gap between the worst $P \in \{2, 4, 8\}$ and the oracle $P_\star$, as well as the
improvement over the $P{=}1$ baseline. The moderate fall-off (mean $-$7.2\% vs.\ oracle) and consistent
accretive gains (mean $+$6.4\% vs.\ baseline) show that even at worst-case $P$, ND-LoRA still beats the
parameter-matched baseline on every hallucination benchmark, mitigating the practical concern of
perfect task-dependent $P$ selection.

\begin{table}[h]
  \centering
  \begin{tabular}{l|cc|rr}
    \textbf{Task} & \textbf{$P^\star$} & \textbf{Worst $P$} & \textbf{$\Delta$Oracle} & \textbf{$\Delta$Baseline} \\
    \hline
    HE-Dialog & 4 & 8 & $-$8.7\% & +3.9\% \\
    HE-QA & 4 & 8 & $-$17.8\% & +1.4\% \\
    HE-Summ & 4 & 8 & $-$2.9\% & +21.9\% \\
    MemoTrap v2 & 8 & 4 & $-$6.2\% & +2.0\% \\
    TF-MC1 & 2 & 4 & $-$2.7\% & +4.4\% \\
    TF-MC2 & 2 & 4 & $-$4.6\% & +4.5\% \\
    \hline
    \textbf{Mean} & & & \textbf{$-$7.2\%} & \textbf{+6.4\%} \\
  \end{tabular}
  \caption{Per-task sensitivity to suboptimal $P$. $P^\star$: task-optimal value. Worst $P$: worst-performing
  value in $\{2, 4, 8\}$. $\Delta$Oracle: gap from $P^\star$. $\Delta$Baseline: improvement over $P{=}1$ baseline.}
  \label{tab:sensitivity}
\end{table}

\subsection{An Interpretable Router For Optimal Number of Streams}
\label{sec:router_model}

To demonstrate that the task-optimal $P_\star$ patterns in \autoref{tab:taskdiv} reflect real structure rather than
arbitrary variation, we train a simple interpretable router that predicts optimal $P_\star$ from prompt features
alone. While more complex routers could improve performance, we prioritize simplicity and interpretability to
understand the underlying structure.

We fit a simple regression on two features, trained on just 10 samples per task with oracle $P$ labels:

\begin{equation}
  \hat{P} = \text{clip}\bigl(
    0.196 \log W - 2.283 Q + 3.321
  \bigr)
  \label{eq:router}
\end{equation}

where $Q$ is the ratio of interrogative to declarative sentences, $W$ measures prompt length in words, and
$\text{clip}(\cdot)$ snaps predictions to the nearest valid $P \in \{1,2,4,8\}$. This two-feature router
achieves 99\% of oracle hallucination performance on held-out samples (97\% across all 12 tasks).

The learned coefficients reveal an interpretable trade-off between \emph{knowledge retrieval} and
\emph{verifiability}. The negative weight on interrogative sentence ratio indicates that question-dense
prompts --- where success depends on precise recall of stored knowledge --- benefit from lower $P$ values
that maximize focus from a single stream. Conversely, the positive weight on word count reflects that
longer prompts --- where success depends on cross-checking claims against provided context --- require
higher $P$ for diverse verification across streams. More broadly, tasks prioritizing retrieval favor low
diversity, while tasks prioritizing verifiability favor high diversity.

\subsection{LoRA Hyperparameter Sensitivity Analysis}
\label{sec:lora_ablations}
A natural concern is whether ND-LoRA's improvements stem from LoRA hyperparameter choices rather than neural
diversity \textit{per se}. We consider three potential confounds: (i) \textit{expressivity}: $P$ parallel
rank-$R$ adapters yield $P \times R$ total parameters, so improvements might reflect capacity rather than
diversity; (ii) \textit{alpha scaling}: different $\alpha/r$ ratios affect update magnitudes and could
change which solutions are reachable; and (iii) \textit{optimization dynamics}: higher-rank adapters might
converge to different basins.

\textbf{Expressivity.} This confound is addressed by parameter matching in the main text
(\autoref{tab:relacc}): baselines use higher-rank single LoRA (R32--R128) to match ND-LoRA's total parameter
count, yet ND-LoRA still outperforms on hallucination benchmarks.

\textbf{Alpha scaling.} We conducted a sensitivity analysis varying single-LoRA rank from $R16$ to $R128$ under two
alpha strategies: constant scaling ($\alpha/r=2$) and constant alpha ($\alpha=32$). Results in
\autoref{tab:lora_constant_scaling} show that constant-scaling single-LoRA is not a suitable baseline for two
reasons. First, the only monotonic trend observed is \textit{degradation} of general capabilities: Wikitext
perplexity increases from 0.776 to 0.795 bits per byte (+2.4\%), TriviaQA-8 drops from 19\% to
17\% (-11\%), and NQ-8 drops from 7\% to 5\% (-29\%) as rank increases from $R16$ to $R128$. Second,
hallucination benchmark performance is unstable across this $8\times$ rank variation: while some pairwise
differences are statistically significant, they're unstable across both rank and benchmarks (e.g. HE-Dialog
vs. HE-QA within R64). We therefore use fixed $\alpha=32$ baselines, which provide stable reference points without the
capability degradation observed under constant scaling. Importantly, ND-LoRA remains statistically
significantly better than both baseline types --- all winners stay winners --- and using constant-scaling
baselines would in fact create additional ND-LoRA wins (e.g. Wikitext BPB, NQ-8 and Winogrande $P=8$).

\begin{table}[htbp]
  \centering
  \begin{tabular}{lcccc}
    Metric & R16 & R32 & R64 & R128 \\
    [-0.75em]\\ \hline \\[-0.75em]
    HE Dialog & 0.46\small{$\pm$0.01} & 0.46\small{$\pm$0.01} & 0.49\small{$\pm$0.01} & 0.45\small{$\pm$0.01} \\
    HE QA & 0.37\small{$\pm$0.01} & 0.37\small{$\pm$0.01} & 0.34\small{$\pm$0.01} & 0.36\small{$\pm$0.01} \\
    HE Summ & 0.41\small{$\pm$0.01} & 0.46\small{$\pm$0.01} & 0.48\small{$\pm$0.01} & 0.41\small{$\pm$0.01} \\
    MemoTrap & 0.64\small{$\pm$0.03} & 0.63\small{$\pm$0.03} & 0.63\small{$\pm$0.03} & 0.64\small{$\pm$0.03} \\
    TF-MC1 & 0.25\small{$\pm$0.03} & 0.25\small{$\pm$0.03} & 0.24\small{$\pm$0.03} & 0.24\small{$\pm$0.03} \\
    TF-MC2 & 0.41\small{$\pm$0.03} & 0.40\small{$\pm$0.03} & 0.39\small{$\pm$0.03} & 0.40\small{$\pm$0.03} \\
    [-0.75em]\\ \hline \\[-0.75em]
    NQ-8 & 0.07\small{$\pm$0.01} & 0.06\small{$\pm$0.01} & 0.06\small{$\pm$0.01} & 0.05\small{$\pm$0.01} \\
    NQ-swap & 0.55\small{$\pm$0.01} & 0.55\small{$\pm$0.01} & 0.55\small{$\pm$0.01} & 0.54\small{$\pm$0.01} \\
    PopQA & 0.11\small{$\pm$0.01} & 0.11\small{$\pm$0.01} & 0.11\small{$\pm$0.01} & 0.11\small{$\pm$0.01} \\
    TQA-8 & 0.19\small{$\pm$0.01} & 0.18\small{$\pm$0.01} & 0.18\small{$\pm$0.01} & 0.17\small{$\pm$0.01} \\
    Wikitext BPB & 0.776 & 0.781 & 0.790 & 0.795 \\
    Winogrande & 0.56\small{$\pm$0.03} & 0.57\small{$\pm$0.03} & 0.58\small{$\pm$0.03} & 0.56\small{$\pm$0.03} \\
    [-0.75em]\\ \hline \\[-0.75em]
    $\alpha/r$ & 2.00 & 2.00 & 2.00 & 2.00 \\
    [-0.75em]\\ \hline \\[-0.75em]
  \end{tabular}
  \caption{
    Constant scaling $\alpha/r=2$: $\alpha$ varies with rank. Hallucination metrics are noisy but many
    general-capability metrics degrade monotonically (e.g. Wikitext BPB 0.776 $\rightarrow$ 0.795, NQ-8 0.07
    $\rightarrow$ 0.05), making this an unsuitable baseline.
  }
  \label{tab:lora_constant_scaling}
\end{table}

\begin{table}[htbp]
  \centering
  \begin{tabular}{lcccc}
    Metric & R16 & R32 & R64 & R128 \\
    [-0.75em]\\ \hline \\[-0.75em]
    HE Dialog & 0.46\small{$\pm$0.01} & 0.46\small{$\pm$0.01} & 0.46\small{$\pm$0.01} & 0.46\small{$\pm$0.01} \\
    HE QA & 0.37\small{$\pm$0.01} & 0.37\small{$\pm$0.01} & 0.34\small{$\pm$0.01} & 0.34\small{$\pm$0.01} \\
    HE Summ & 0.41\small{$\pm$0.01} & 0.40\small{$\pm$0.01} & 0.39\small{$\pm$0.01} & 0.38\small{$\pm$0.01} \\
    MemoTrap & 0.64\small{$\pm$0.03} & 0.63\small{$\pm$0.03} & 0.63\small{$\pm$0.03} & 0.63\small{$\pm$0.03} \\
    TF-MC1 & 0.25\small{$\pm$0.03} & 0.25\small{$\pm$0.03} & 0.24\small{$\pm$0.03} & 0.25\small{$\pm$0.03} \\
    TF-MC2 & 0.41\small{$\pm$0.03} & 0.40\small{$\pm$0.03} & 0.40\small{$\pm$0.03} & 0.41\small{$\pm$0.03} \\
    [-0.75em]\\ \hline \\[-0.75em]
    NQ-8 & 0.07\small{$\pm$0.01} & 0.07\small{$\pm$0.01} & 0.06\small{$\pm$0.01} & 0.07\small{$\pm$0.01} \\
    NQ-swap & 0.55\small{$\pm$0.01} & 0.55\small{$\pm$0.01} & 0.55\small{$\pm$0.01} & 0.55\small{$\pm$0.01} \\
    PopQA & 0.11\small{$\pm$0.01} & 0.11\small{$\pm$0.01} & 0.11\small{$\pm$0.01} & 0.11\small{$\pm$0.01} \\
    TQA-8 & 0.19\small{$\pm$0.01} & 0.19\small{$\pm$0.01} & 0.19\small{$\pm$0.01} & 0.19\small{$\pm$0.01} \\
    Wikitext BPB & 0.776 & 0.775 & 0.778 & 0.778 \\
    Winogrande & 0.56\small{$\pm$0.03} & 0.57\small{$\pm$0.03} & 0.56\small{$\pm$0.03} & 0.57\small{$\pm$0.03} \\
    [-0.75em]\\ \hline \\[-0.75em]
    $\alpha/r$ & 2.00 & 1.00 & 0.50 & 0.25 \\
    [-0.75em]\\ \hline \\[-0.75em]
  \end{tabular}
  \caption{Constant $\alpha=32$: scaling varies with rank. Most metrics are stable alongside general
  capabilities, helping rule out expressivity and optimization dynamics as confounds.}
  \label{tab:lora_constant_alpha}
\end{table}

\textbf{Adapter Modules.} To characterize which LoRA targets contribute to hallucination reduction, we ablate modules
from the full StreamLoRA-BT configuration (LoRA on both attention and MLP projections), which serves as the reference
point. Two variants each keep LoRA on one module family only:
\begin{itemize}
  \item \emph{Attention-Only}: LoRA applied to attention projections (\texttt{q\_proj}, \texttt{k\_proj},
    \texttt{v\_proj}, \texttt{o\_proj}) only; MLP projections left unadapted
  \item \emph{MLP-Only}: LoRA applied to MLP projections (\texttt{gate\_proj}, \texttt{up\_proj},
    \texttt{down\_proj}) only; attention projections left unadapted
\end{itemize}

\begin{table}[htbp]
  \centering
  \begin{tabular}{l|cc}
    \textbf{Task} & \textbf{$\Delta$\% Attention-Only} & \textbf{$\Delta$\% MLP-Only} \\
    \hline
    HaluEval Dialog & -1.7\% & -0.6\% \\
    HaluEval QA & +16.8\% & -1.8\% \\
    HaluEval Summarization & -5.3\% & -27.0\% \\
    MemoTrap v2 & +2.5\% & +0.9\% \\
    NQ (8-shot) & +11.7\% & -1.7\% \\
    PopQA & -0.8\% & -0.8\% \\
    TriviaQA (8-shot) & -5.0\% & -6.9\% \\
    TruthfulQA MC1 & +3.1\% & +2.4\% \\
    TruthfulQA MC2 & +0.2\% & +1.4\% \\
  \end{tabular}
  \caption{
    LoRA module ablation results. Columns are relative percentage changes from the full StreamLoRA-BT
    configuration (the reference point). Evaluations performed on N=1024 samples per task.
  }
  \label{tab:lora_ablations_delta_pct}
\end{table}

\textbf{Optimization dynamics.} Under fixed alpha ($\alpha=32$), general capabilities remain stable across
$8\times$ rank variation: Wikitext BPB is flat (0.775--0.778) and Winogrande accuracy is statistically
indistinguishable (0.56--0.57) across R16--R128 (\autoref{tab:lora_constant_alpha}). If optimization dynamics
differed meaningfully across rank (e.g. higher-rank adapters converging to different loss basins) we would
expect divergence on these general capability metrics. The observed stability indicates that fixed-alpha
configurations converge to similar solutions regardless of rank, ruling out optimization dynamics as a
confound for ND-LoRA's hallucination improvements.

\subsection{Regularization Hyperparameter Sensitivity Analysis}
\label{sec:reg_sensitivity}

The main experiments fix two regularization hyperparameters: the Barlow Twins loss weight
$\lambda_\text{BT}$ and the design layer $\ell_\star$ at which decorrelation is applied.
To assess sensitivity, we sweep over $\lambda_\text{BT} \in [0.01, 0.50]$ (log-scale) and
$\ell_\star \in [5, 23]$ at $P=2$, evaluating a subset of benchmarks on 51 non-divergent configurations and
summarize the results below.

\autoref{tab:design_layer_sensitivity} reports design layer sensitivity.
Per-task hallucination scores are stable across layers 7--23, with no layer bin differing by more than
2 points on any benchmark. No single layer is critical for the method's effectiveness.

\begin{table}[htbp]
  \centering
  \begin{tabular}{lcccccc}
    Metric & 7--9 & 10--12 & 13--15 & 16--18 & 19--21 & 22--23 \\
    [-0.75em]\\ \hline \\[-0.75em]
    HE-Dial & 0.459 & 0.444 & 0.455 & 0.443 & 0.451 & 0.450 \\
    HE-QA & 0.350 & 0.345 & 0.346 & 0.347 & 0.351 & 0.345 \\
    HE-Summ & 0.498 & 0.479 & 0.481 & 0.470 & 0.474 & 0.451 \\
    MemoTrap & 0.650 & 0.663 & 0.666 & 0.662 & 0.666 & 0.657 \\
    TF-MC2 & 0.389 & 0.387 & 0.393 & 0.390 & 0.402 & 0.391 \\
    [-0.75em]\\ \hline \\[-0.75em]
    Wikitext BPB & 0.800 & 0.793 & 0.795 & 0.795 & 0.794 & 0.789 \\
    [-0.75em]\\ \hline \\[-0.75em]
  \end{tabular}
  \caption{Design layer sensitivity ($P=2$). Columns are design layer
    $\ell_\star$ bins. Hallucination metrics are stable across layers 7--23;
  no single layer is critical for the method's effectiveness.}
  \label{tab:design_layer_sensitivity}
\end{table}

\autoref{tab:lambda_bt_sensitivity} reports $\lambda_\text{BT}$ sensitivity.
Stronger regularization improves most hallucination benchmarks (e.g.\ HE-QA: 0.320 $\rightarrow$ 0.404,
MemoTrap: 0.656 $\rightarrow$ 0.678) but degrades perplexity (0.786 $\rightarrow$ 0.821 BPB),
reflecting a hallucination--perplexity tradeoff.

\begin{table}[htbp]
  \centering
  \begin{tabular}{lccccc}
    Metric & 0.01--0.03 & 0.03--0.10 & 0.10--0.20 & 0.20--0.35 & 0.35--0.50 \\
    [-0.75em]\\ \hline \\[-0.75em]
    HE-Dial & 0.437 & 0.432 & 0.454 & 0.475 & 0.467 \\
    HE-QA & 0.320 & 0.332 & 0.364 & 0.365 & 0.398 \\
    HE-Summ & 0.468 & 0.448 & 0.494 & 0.489 & 0.464 \\
    MemoTrap & 0.656 & 0.661 & 0.666 & 0.667 & 0.673 \\
    TF-MC2 & 0.387 & 0.393 & 0.393 & 0.400 & 0.393 \\
    [-0.75em]\\ \hline \\[-0.75em]
    Wikitext BPB & 0.786 & 0.786 & 0.796 & 0.802 & 0.814 \\
    [-0.75em]\\ \hline \\[-0.75em]
  \end{tabular}
  \caption{$\lambda_\text{BT}$ sensitivity ($P=2$). Columns are
    $\lambda_\text{BT}$ bins. Stronger regularization improves hallucination metrics
  but degrades perplexity, reflecting a hallucination--perplexity tradeoff.}
  \label{tab:lambda_bt_sensitivity}
\end{table}

Together with the LoRA rank analysis above (\autoref{sec:lora_ablations}),
the $P \times$ method grid (\autoref{sec:complete_results}), the LoRA module ablation
(\autoref{sec:lora_ablations}), and the per-task $P$-sensitivity (\autoref{sec:sensitivity}), we evaluate
over 70 total configurations spanning LoRA rank ($R16$--$R128$), $P$ ($\{1,2,4,8\}$), $\lambda_\text{BT}$
($[0.01, 0.50]$), design layer ($[7, 23]$), and LoRA module targets.

\subsection{Glossary}
\label{sec:glossary}
Our analysis combines terminology from deep learning, financial econometrics, and high-dimensional
probability. For readers unfamiliar with any of these fields, we provide brief intuitions for the
non-standard terms used throughout the paper.

\begin{description}
  \item[Portfolio theory] A framework from financial economics \citep{markowitz1952portfolio} for
    constructing collections of risky assets that minimize total risk (variance) for a given expected return.
    The central insight is that combining assets with low pairwise correlation reduces portfolio variance
    below that of any individual asset --- diversification. We adapt this framework by treating each parallel
    stream's output as an asset whose ``risk'' is its deviation from the oracle output.
  \item[Portfolio-theoretic diversification] The act of reducing risk by holding multiple uncorrelated
    positions rather than concentrating exposure in one. In our setting, running multiple decorrelated
    streams reduces the variance of the aggregated output and therefore the probability of a catastrophic hallucination.
  \item[Second-moment reliability] Performance guarantees based on the variance and covariance structure of
    errors (second-order statistics), rather than on expected loss (first-moment / mean performance).
    Classical ensemble learning targets mean error; we target tail probability, which is controlled by second
    moments via Chebyshev-type inequalities.
  \item[Tail risk] The probability of rare but catastrophic failures --- i.e. the tail of the error
    distribution. A model can have excellent average performance yet unacceptable tail risk.
  \item[Signal-to-noise ratio (SNR)] The ratio of signal magnitude to noise magnitude; here defined as $SNR
    \triangleq \delta^2 / \bar{\sigma}^2$, where $\delta$ is the tolerance and $\bar{\sigma}^2$ is the
    average stream noise variance. Higher SNR means hallucinations are easier to avoid.
  \item[Kurtosis bound ($C_4$)] A constant controlling the heaviness of the noise distribution's tails via a
    finite fourth moment: $\E[\|\xi_i\|_2^4] \le C_4 \sigma_i^4$. This is a mild assumption that excludes
    only pathological distributions with infinite fourth moments.
  \item[Condition number ($\kappa$)] For a matrix $A$, the ratio $s_{\max}/s_{\min}$ of its largest
    to smallest singular value. It measures how much $A$ can stretch some directions relative to others;
    $\kappa = 1$ is isotropic, $\kappa \gg 1$ is ill-conditioned. Here it bounds how much the readout $A$ can
    distort cosine similarities between representations.
  \item[Spectral bound] A bound on a quantity in terms of the eigenvalues or singular values of an associated
    matrix. We use spectral bounds on $A^\top A$ to control distortion of inner products through the readout.
  \item[Norm concentration] The phenomenon that in high dimensions, random vectors of interest concentrate
    near a sphere of predictable radius (typically $\sqrt{d}$) with small relative variance. Standard in
    high-dimensional probability \citep{vershynin2018high}.
  \item[Lipschitz decoding] An assumption that small changes in the hidden representation produce
    proportionally small changes in the output: $\|f(z) - f(z')\|_2 \le L\|z - z'\|_2$. Standard in neural
    network analysis \citep{fazlyab2019efficient, bartlett2017spectrally}.
  \item[Whitening] The preprocessing step of transforming features to have zero mean and identity covariance.
    Standard in self-supervised learning and used in our definition of $\tilde{z}_i$.
\end{description}

\subsection{Notation}
\label{sec:notation}
\autoref{tab:notation} summarizes all mathematical notation used throughout the paper.

\begin{longtable}{llp{0.58\textwidth}}
  \caption{Summary of notation.} \label{tab:notation} \\
  \hline
  \textbf{Symbol} & \textbf{Domain} & \textbf{Meaning} \\
  \hline
  \endfirsthead
  \hline
  \textbf{Symbol} & \textbf{Domain} & \textbf{Meaning} \\
  \hline
  \endhead
  \hline
  \endfoot
  $A$ & $\mathbb{R}^{V \times d}$ & Shared local-linear readout: $\xi_i = A \tilde{z}_i$. \\
  $B$ & $\mathbb{N}$ & Batch size. \\
  $\mathcal{B}(P)$ & $[0,1]$ & Hallucination bound: $\mathcal{B}(P) \triangleq v(P)/(v(P) + \delta^2)$. \\
  $C^{(ij)}$ & $\mathbb{R}^{d \times d}$ & Cross-correlation matrix: $C^{(ij)} \triangleq \E[\tilde{z}_i
  \tilde{z}_j^\top]$. \\
  $C_4$ & $[1, \infty)$ & Kurtosis bound: $\E[\|\xi_i\|_2^4] \le C_4 \sigma_i^4$. \\
  $C_\ast$ & $\mathbb{R}_{>0}$ & Readout-kurtosis constant: $C_\ast \triangleq \sqrt{C_4}\,\kappa^2$. \\
  $\mathcal{D}$ & $[0, 1]$ & Neural diversity index (\autoref{eq:ndi}); $0$ orthogonal, $1$ collapsed. \\
  $\mathcal{D}_{ij}$ & $[0, 1]$ & Cosine similarity between streams $i, j$ in representation space. \\
  $\mathcal{D}_\xi, \mathcal{D}_{\xi,ij}$ & $[0, 1]$ & Analogous quantities in readout (noise) space. \\
  $d$ & $\mathbb{N}$ & Hidden dimension. \\
  $E_w$ & $\mathbb{R}_{\ge 0}$ & Output error: $E_w \triangleq \|\widehat{Y}_w(x) - y_\star(x)\|_F$. \\
  $\{e_k\}$ & $\mathbb{R}^d$ & Right singular vectors of $A$ (eigenvectors of $A^\top A$). \\
  $f$ & $\mathbb{R}^d \to \mathbb{R}^V$ & Decoding function: $\widehat{Y}_w(x) = f(\widehat{Z}_w(x))$. \\
  $\HALL$ & event & Hallucination event: $\HALL \triangleq \{E_w \ge \delta\}$. \\
  $i, j$ & $\{1, \ldots, P\}$ & Stream indices. \\
  $L$ & $\mathbb{R}_{>0}$ & Lipschitz constant of $f$. \\
  $\ell_\star$ & $\mathbb{N}$ & Pre-specified design layer at which decorrelation is applied. \\
  $\mathcal{L}_{BT}$ & $\mathbb{R}_{\ge 0}$ & Barlow Twins loss: $\mathcal{L}_{BT} \triangleq
  \E_{i<j}\|C^{(ij)} - I\|_F$. \\
  $\mathcal{L}_{CE}$ & $\mathbb{R}_{\ge 0}$ & Cross-entropy loss. \\
  $P$ & $\mathbb{N}$ & Number of parallel computational streams. \\
  $P_\star$ & $\mathbb{N}$ & Optimal number of streams minimizing $\Prob(\HALL)$. \\
  $\Prob(\HALL)$ & $[0, 1]$ & Hallucination probability. \\
  $SNR$ & $\mathbb{R}_{>0}$ & Signal-to-noise ratio: $SNR \triangleq \delta^2 / \bar{\sigma}^2$. \\
  $T$ & $\mathbb{N}$ & Sequence length. \\
  $V$ & $\mathbb{N}$ & Vocabulary / output dimension. \\
  $s_k, s_{\min}, s_{\max}$ & $\mathbb{R}_{>0}$ & Singular values of $A$; $s_{\min} \le s_k \le s_{\max}$. \\
  $v(P)$ & $\mathbb{R}_{\ge 0}$ & Variance of aggregated error: $v(P) \triangleq \Var(E_w)$. \\
  $w_i$ & $[0, 1]$ & Aggregation weight for stream $i$; $\sum_i w_i = 1$. \\
  $X$ & set & Input space. \\
  $x$ & $X$ & Input query. \\
  $\widehat{Y}_w(x)$ & $\mathbb{R}^V$ & Aggregated prediction. \\
  $y_\star(x)$ & $\mathbb{R}^V$ & Oracle (ground-truth) output. \\
  $Z_i$ & $\mathbb{R}^d$ & Hidden output of stream $i$: $Z_i = z_\star + \varepsilon_i$. \\
  $\widehat{Z}_w$ & $\mathbb{R}^d$ & Aggregated hidden representation: $\widehat{Z}_w \triangleq \sum_i w_i Z_i$. \\
  $\tilde{z}_i$ & $\mathbb{R}^d$ & Per-feature whitened stream representation. \\
  $z_\star(x)$ & $\mathbb{R}^d$ & Oracle hidden representation. \\
  $\beta, \gamma$ & $\mathbb{R}_{>0}$ & Growth parameters in $\bar{\rho}(P) = \rho_0 + \beta(P-1)^\gamma$. \\
  $\delta$ & $\mathbb{R}_{>0}$ & Error tolerance. \\
  $\varepsilon_i$ & $\mathbb{R}^d$ & Centered stream noise with variance $\sigma_i^2$. \\
  $\kappa$ & $[1, \infty)$ & Condition number of $A$: $\kappa \triangleq s_{\max}/s_{\min}$. \\
  $\lambda_{BT}$ & $\mathbb{R}_{\ge 0}$ & Weight on the Barlow Twins term in the total loss. \\
  $\rho_0$ & $[0, 1)$ & Base correlation in $\bar{\rho}(P) = \rho_0 + \beta(P-1)^\gamma$. \\
  $\rho_{ij}$ & $[-1, 1]$ & Pairwise noise correlation between streams $i \ne j$. \\
  $\bar{\rho}$ & $[-1, 1]$ & Average pairwise noise correlation: $\bar{\rho} \triangleq \E_{i<j}[\rho_{ij}]$. \\
  $\bar{\rho}(P)$ & $[0, 1]$ & Correlation growth model as a function of $P$. \\
  $\sigma_i^2$ & $\mathbb{R}_{>0}$ & Noise variance of stream $i$. \\
  $\bar{\sigma}^2$ & $\mathbb{R}_{>0}$ & Average per-stream noise variance: $\bar{\sigma}^2 \triangleq
  \E[\sigma_i^2]$. \\
  $\Sigma$ & $\mathbb{R}^{P \times P}$ & Noise covariance matrix. \\
  $\xi_i$ & $\mathbb{R}^d$ & Noise vector of stream $i$ in readout space: $\xi_i = A \tilde{z}_i$. \\
\end{longtable}

\end{document}

%% file: math_commands.tex
\usepackage{amsmath,amsfonts,bm}

\def\eqref#1{equation~\ref{#1}}

\def\1{\bm{1}}

\DeclareMathAlphabet{\mathsfit}{\encodingdefault}{\sfdefault}{m}{sl}
\SetMathAlphabet{\mathsfit}{bold}{\encodingdefault}{\sfdefault}{bx}{n}

\DeclareMathOperator*{\E}{\mathbb{E}}

\newcommand{\Var}{\mathrm{Var}}

\newcommand{\Corr}{\mathrm{Corr}}

